\documentclass[conference]{IEEEtran}
\IEEEoverridecommandlockouts

\pdfoutput=1

\usepackage[longend,ruled,vlined,linesnumbered]{algorithm2e}

\usepackage{enumitem}

\usepackage{microtype}                 
\usepackage{textcomp}                  
\usepackage{times}                     
\usepackage{tabu}                      
\usepackage{booktabs}                  

\usepackage{graphicx}
\usepackage{color}
\usepackage{subfigure}
\usepackage{url,times,amsmath,cases, amssymb}
\usepackage{diagbox}
\usepackage{multirow}
\usepackage{booktabs}
\usepackage{caption}

\usepackage{amssymb} 
\usepackage[normalem]{ulem}
\usepackage{hyperref}
\usepackage{booktabs}
\usepackage{xparse}
\ExplSyntaxOn
\NewDocumentCommand{\longdash}{ O{2} }
 {
  --\prg_replicate:nn { #1 - 1 } { \negthinspace -- }
 }
\ExplSyntaxOff

\newcommand{\Break}{\textbf{Break}}

\def\_{\rule{.3em}{.15ex}}      

\newtheorem{definition}{Definition}
\newtheorem{property}{Property}
\newtheorem{proposition}{Observation}
\newtheorem{lemma}{Lemma}
\newtheorem{theorem}{Theorem}
\newtheorem{corollary}{Corollary}

\newcommand {\mymarginpar}[1]{\marginpar{#1}}
\renewcommand {\marginpar}[1]{}

\newcommand {\rfig}[1]{Figure \ref{fig:#1}}

\newcommand {\bsec}[2]{\section{#1}
                       \label{sec:#2} }

\newcommand {\rsec}[1]{Section \ref{sec:#1}}

\newcommand {\bsubsec}[2]{\mymarginpar{sec:#2}
                       \subsection{#1}
                       \label{sec:#2} }

\newcommand {\rsubsec}[1]{Section \ref{sec:#1}}

\newcommand {\rsuba}[1]{Appendix \ref{sec:#1}}

\newcommand {\beq}[1]{
                      \begin{equation}
                      \label{eq:#1} }
\newcommand {\eeq}{\end{equation}}
\newcommand {\beqno}[1]{\begin{eqnarray}
                      \nonumber}

\newcommand {\eeqno}{ && \end{eqnarray}}
\newcommand {\req}[1]{Eq.~(\ref{eq:#1})}

\newcommand {\bear}[1]{
                       \begin{eqnarray}
                       \label{eq:#1} }

\newcommand {\bearno}[1]{
                       \begin{eqnarray}
                       \nonumber}

\newcommand {\eear}{\end{eqnarray}}
\newcommand {\eearno}{\end{eqnarray}}

\newcommand {\btab}[1]{
                       \begin{table}
                       \centering
                       \begin{tabular}{#1}}
\newcommand {\etab}[3] {
                       \end{tabular}
                       \caption[#3]{#2}
                       \label{tab:#1}
                       \end{table}
                       \vspace{.1in}}
\newcommand {\rtab}[1]{Table \ref{tab:#1}}

\newcommand {\btabular}[1]{\begin{center}
                       \begin{tabular}{#1}}
\newcommand {\etabular}{\end{tabular}
                       \end{center}}

\newcommand {\bdefin}[1]{\begin{definition}\label{def:#1}}
\newcommand {\edefin}       {\end{definition}}
\newcommand {\rdef}[1]{Definition \ref{def:#1}}

\newcommand {\bpro}[1]{\begin{property}
                      \label{pro:#1} }
\newcommand {\epro}   {\end{property}}

\newcommand {\bprop}[1]{\begin{proposition}
                      \label{prop:#1} }
\newcommand {\eprop}       {\end{proposition}}
\newcommand {\rprop}[1]{Observation \ref{prop:#1}}

\newcommand {\blem}[1]{\begin{lemma}
                      \label{lem:#1}}
\newcommand {\elem}   {\end{lemma}}

\newcommand {\bthe}[1]{\begin{theorem}
                      \label{the:#1} }
\newcommand {\ethe}   {\end{theorem}}
\newcommand {\rthe}[1]{Theorem \ref{the:#1}}

\newcommand {\bproof}{\noindent {\bf Proof.} \ }
\newcommand {\eproof} {$\hfill \blacksquare$ \\ \vspace{.3cm}}
\newcommand {\bcor}[1]{\begin{corollary}
                      \label{cor:#1} }
\newcommand {\ecor}   {\end{corollary}}

\newcommand {\ralg}[1]{Algorithm \ref{alg:#1}}
\newcommand{\hide}[1]{}

\def\BibTeX{{\rm B\kern-.05em{\sc i\kern-.025em b}\kern-.08em
    T\kern-.1667em\lower.7ex\hbox{E}\kern-.125emX}}
\begin{document}

\title{\huge Mobility Inference on Long-Tailed Sparse Trajectory
}

\author{\IEEEauthorblockN{Lei Shi}
\IEEEauthorblockA{\textit{Department of Computer Science and Engineering} \\
\textit{Beihang University}\\
Beijing, China \\
leishi@buaa.edu.cn}
}

\maketitle

\begin{abstract}
Analyzing the urban trajectory in cities has become an important topic in data mining. How can we model the human mobility consisting of stay and travel from the raw trajectory data? How can we infer such a mobility model from the single trajectory information? How can we further generalize the mobility inference to accommodate the real-world trajectory data that is sparsely sampled over time?

In this paper, based on formal and rigid definitions of the stay/travel mobility, we propose a single trajectory inference algorithm that utilizes a generic long-tailed sparsity pattern in the large-scale trajectory data. The algorithm guarantees a 100\% precision in the stay/travel inference with a provable lower-bound in the recall. Furthermore, we introduce an encoder-decoder learning architecture that admits multiple trajectories as inputs. The architecture is optimized for the mobility inference problem through customized embedding and learning mechanism. Evaluations with three trajectory data sets of 40 million urban users validate the performance guarantees of the proposed inference algorithm and demonstrate the superiority of our deep learning model, in comparison to well-known sequence learning methods. On extremely sparse trajectories, the deep learning model achieves a 2$\times$ overall accuracy improvement from the single trajectory inference algorithm, through proven scalability and generalizability to large-scale versatile training data.
\end{abstract}


\bsec{Introduction}{Intro}




The recent surge of metropolitan-scale human trajectory data, e.g., mobile traces \cite{Ratti06}, taxi logs \cite{yuan2013t}, and geo-referenced check-ins \cite{GeoTweet13}, paves the way for a fundamental understanding of the human mobility in cities.
In both theoretical and empirical studies, the urban trajectory of human is considered as interleaving segments of {\tt stay} and {\tt travel} \cite{gonzalez2008understanding}\cite{brockmann2006scaling}\cite{calabrese2010geography}. The inference of these segments from the raw trajectory data plays a pivotal role in solving many urban analytics tasks. For example, in traffic planning and optimization, the detected travels are used as the training data for the travel time estimation \cite{wang2014travel}\cite{Thiagarajan:2009:VAE}. In trade area analysis, the discovery of user's visits to business sites relies on the segmentation of stays and travels from the trajectory data \cite{qu2013trade}.



In the literature, there is a consensus that the stay segments (also known as the stops) can be defined as the part of the trajectory within a spatially constrained region for a sufficiently long time \cite{calabrese2010geography}\cite{Phithakkitnukoon10}\cite{jiang2013review}. Algorithms have been proposed that first partition the trajectory at all the record intervals larger than a spatial threshold and infer the resulting sub-trajectories as stay by the definition \cite{calabrese2010geography}\cite{Phithakkitnukoon10}. On the other hand, no definition of the travel segment has been formulated on the trajectory data. Existing works mostly assume a dense sampling rate in the trajectory data, i.e., seconds or a few minutes on average between the consecutive trajectory records \cite{herring2010estimating}\cite{rahmani2013path}\cite{li2017citywide}. For such data, the real-time speed of the trajectory can be calculated, which is used to accurately detect all the travels.

\begin{figure}[t]
\centering
\includegraphics[width=3.4 in]{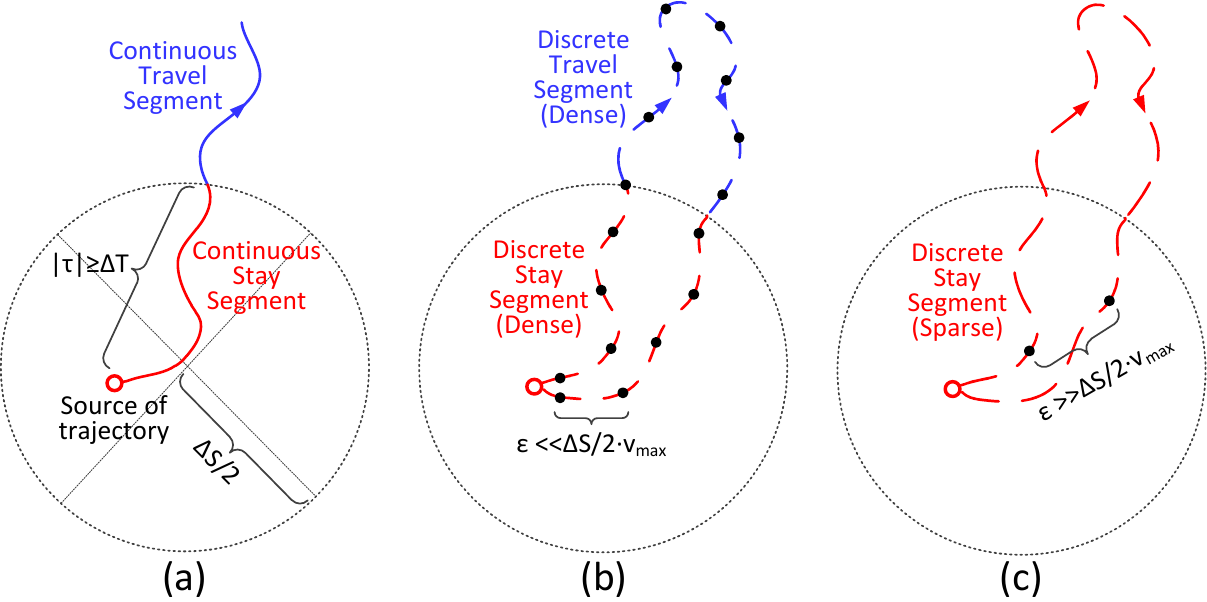}
\vspace{-0.07 in}
\caption{Illustrative examples of \rdef{Continuous-Segment} and \rdef{Discrete-Segment}: (a) continuous stay/travel segments; discrete stay/travel segments on (b) dense trajectory; (c) sparse trajectory.}
\vspace{-0.15 in}
\label{fig:Definition}
\end{figure}

Nevertheless, the metropolitan-scale measurement of human trajectories is often extremely sparse over time for pragmatic constraints such as the power consumption and the user privacy. In the mobile sensing data used in this paper, the average record interval is as long as two hours, two magnitudes larger than that of the previously considered trajectory data. The existing mobility inference algorithms designed for the dense trajectories do not work any more. For instance, two consecutive records in a trajectory with a 24-hour interval reported at the nearby locations will be identified as in the same stay segment (\rfig{Definition}(c)). In fact, these records could be either the separate stays at home or two pass-bys during the daily commute. We are {\em agnostic} about their mobility given the single trajectory information only.



The inference of stay and travel on sparsely sampled trajectories are highly challenging. First, the real-world human movement is a complex process with varied speeds and spatiotemporal patterns. 
How can we have a comprehensive definition of stay and travel on the human trajectory for various applications? Second, with the mobility definition, how do we know which part of the trajectory can be inferred as stay or travel using the single trajectory information? How can we design the inference algorithm to work with the metropolitan-scale trajectory data with billions of records? Third, it has been known that the human movement exhibits strong regularity (e.g., a 93\% predictability \cite{Science10}). How can we leverage such regularity to overcome the limit of the single trajectory inference?



To answer the aforementioned questions, we make the following contributions in this work.
\begin{itemize}
 \item \textbf{The formal definition} of both stay and travel on the continuous trajectory model using a pair of spatial and temporal parameters. The linkage of this continuous mobility model to the sparse trajectory data is rigorously studied, which helps to formulate our research problem. (\rsec{Problem})
 \item \textbf{The single trajectory inference algorithm} called Slice \& Doubly Sliding (SDS) designed according to a generic long-tailed sparsity pattern in our trajectory data (\rsec{Analysis}). The algorithm is proved to guarantee a 100\% inference precision and a lower-bounded recall subject to the single trajectory information. (\rsec{Labeling})
 \item \textbf{The optimized encoder-decoder architecture} that captures the regularity of human mobility at the population scale. Several improved architecture designs are introduced to cope with the mobility inference problem, including the decoder mask on unlabeled records, the attention mechanism for extra long trajectories, and the embedding of the mobility-related space and time information. (\rsec{Model})
\end{itemize}
We evaluate the proposed SDS algorithm and the encoder-decoder architecture on both the simulated trajectory data and a sparse trajectory data set characterizing the mobility of 40 million residents in three major Chinese cities (\rsec{Eva}). The experiment results validate the theoretical performance of the SDS algorithm and demonstrate three key advantages of the deep learning model on mobility inference: the capability to utilize the spatiotemporal information of multiple trajectories, the scalability to large training data, and the generalizability to different sets of trajectories.




\bsec{Problem Definition}{Problem}

We first consider the urban trajectory defined by a continuous mobility model. During a time period $T$, a user trajectory $\Gamma$ is composed of a list of temporally continuous records by $\Gamma=\bigcup_{t \in T}<t,\ell(t)>$. $\ell(t)$ denotes the location of a user at time $t$.

\bdefin{Continuous-Segment}\textsc{Continuous mobility of a trajectory} -- On the continuous segment of the trajectory $\Gamma$ during a time period $\tau \subseteq T$, denoted as $\gamma=\bigcup_{t \in \tau}<t,\ell(t)>$, we define its mobility by:

(a) $\gamma$ is a \textbf{stay} segment if: $|\tau| \geq \Delta{T}$ and $||\ell(t_i)-\ell(t_j)||<\Delta{S}$ ($\forall t_i,t_j \in \tau$);

(b) $\gamma$ is a \textbf{travel} segment if: $\gamma$ does not overlap with any continuous segment satisfying (a).
\edefin
Here $|\cdot|$ denotes the length of a time period, $||\cdot||$ is the $L_2$ norm that computes the spatial distance between two records. $\Delta{T}$ and $\Delta{S}$ are the temporal and spatial parameters in the mobility definition.

As shown by the red curve in \rfig{Definition}(a) where the hollow node stands for a long-time stop, \rdef{Continuous-Segment}(a) models the stay segment as a sufficiently long time period ($\geq\Delta{T}$) when the trajectory is kept within a circular region of radius $\Delta{S}/2$. This definition is consistent among all the previous literature \cite{calabrese2010geography}\cite{Phithakkitnukoon10}\cite{jiang2013review}. Note that the stay segments by definition can overlap with each other in space and time. Their enclosure is called the maximal stay segment. On the other hand, based on the ground truth that a user can either stay or travel in any time point, the segment not overlapped with any stays is defined as the travel segment (\rdef{Continuous-Segment}(b)).

The continuous mobility model can not be exactly computed in the real world as the human trajectory is hardly measured continuously. In most cases, the trajectory is composed of a list of discrete records on certain time points (e.g., $t_1<\cdots<t_L$) denoted by $\Gamma = \bigcup_{t \in \{t_1,\cdots,t_L\}}<t,\ell(t)>$ where $L$ denotes the size of the trajectory. A discrete mobility model can be defined in analogy to the continuous model.

\bdefin{Discrete-Segment}\textsc{Discrete mobility of a trajectory} -- On the discrete segment of the trajectory $\Gamma$ in a time series $\omega = \{t_p,\cdots,t_q\}\;(1 \leq p < q \leq L)$, denoted as $\gamma=\bigcup_{t \in \omega}<t,\ell(t)>$, define its mobility by:

(a) $\gamma$ is stay if: $t_q-t_p \geq \Delta{T}$ and $||\ell(t_i)-\ell(t_j)||<\Delta{S}$ ($\forall t_i,t_j \in \omega$);

(b) $\gamma$ is travel if: $\gamma$ does not overlap with any discrete segment satisfying (a).
\edefin

The discrete mobility model can be optimally computed by an exact algorithm (\ralg{Optimal}). Nevertheless, the resulting mobility is not always equivalent to that of the continuous model with the full trajectory information. For example, in \rfig{Definition}(b), the stay and travel segments detected on a densely sampled trajectory by the discrete mobility model generally echo those by the continuous model (\rfig{Definition}(a)). In comparison, the detected segments shown by \rfig{Definition}(c) on the same but sparsely sampled trajectory turn out to be erroneous and largely deviate from the continuous model. The theorem below reveals the relationship between the two models.

\bthe{Continuous-Discrete}\textsc{Intrinsic linkage between discrete and continuous mobility of a trajectory} -- Consider a discrete segment $\gamma$ of the trajectory. Let $\epsilon$ be the maximal time interval between the consecutive records of $\gamma$, $v_{max}$ be the maximal movement speed in $\gamma$:

(a) $\gamma$ satisfying \rdef{Discrete-Segment}(a) under the parameters of $\Delta{S}$ and $\Delta{T}$ is also a stay segment by \rdef{Continuous-Segment}(a) in the continuous model under the parameters of $\Delta'{S} = \Delta{S} + 2 \cdot \epsilon \cdot v_{max}$ and $\Delta'{T} = \Delta{T}$;

(b) $\gamma$ satisfying \rdef{Discrete-Segment}(b) under the parameters of $\Delta{S}$ and $\Delta{T}$ is also a travel segment by \rdef{Continuous-Segment}(b) in the continuous model under the parameters of $\Delta'{S} = \Delta{S}$ and $\Delta'{T} = \Delta{T} + 2 \cdot \epsilon$.
\ethe

The proof is given in \rsuba{Proof}. By \rthe{Continuous-Discrete}, for the discrete trajectory satisfying $\epsilon << \min{(\frac{\Delta{S}}{2 \cdot v_{max}},\frac{\Delta{T}}{2})}$, i.e., having a dense sampling rate, the discrete mobility of the trajectory computed by the exact algorithm can approximate its continuous mobility with tiny parameter changes. Unfortunately, the measurement of human trajectories in big cities is often extremely sparse over time for pragmatic constraints such as the power consumption and the user privacy (e.g., the data set in \rsubsec{Source}).
This work studies the inference of the continuous mobility from the sparse trajectory, which can not be approximated by \rthe{Continuous-Discrete}.

\vspace{.2cm}
\noindent \textsc{\textbf{Problem: Mobility Inference on Sparse Trajectory}}

\noindent \textbf{Given:} (1) \textit{a set of urban users}; (2) \textit{each user's sparse trajectory $\Gamma=\bigcup_{i \in [1,L]}<t_i,\ell(t_i)>$ that $\exists j \in [1,L), ||t_j - t_{j+1}|| > \min(\frac{\Delta{S}}{2 \cdot v_{max}},\frac{\Delta{T}}{2})$}; (3) \textit{the parameters of $\Delta{S}$ and $\Delta{T}$ that define the mobility of the trajectory}.

\noindent \textbf{Infer:} \textit{the continuous mobility of the sparse trajectory at the time of each record, which is denoted by $I_{S/T}(t_i), \forall i \in [1,L]$}. 


Note that the parameters of $\Delta{S}$ and $\Delta{T}$ determine the spatiotemporal scale of mobility. Unless otherwise noted, we use $\Delta{S}=800~m$, $\Delta{T}=30~min$ to study the intra-city mobility. The parameter selection is discussed in \rsuba{Proposition}.

\bsec{Sparsity Analysis on Trajectories}{Analysis}

By \rthe{Continuous-Discrete}, our research problem seems intractable on sparse trajectories. In this section, we analyze a set of real-world trajectory data and discover a generic sparsity pattern that can be utilized in accurately inferring the human mobility.

\bsubsec{Data Source}{Source}




The trajectory data is provided by a mobile analytics company that keeps track of billions of smart devices in China, including mobile phones, tablets, wearable devices, etc. The company's third-party APIs are registered inside more than 100,000 types of mobile apps in a wide spectrum of domains. When a registered app is activated on a device (not necessarily being used), the API will report the location of that device to the company server. The metadata of each trajectory record is shown in \rtab{DataFormat}.


We extract the full-scale trajectory data within three major Chinese cities during a period of 90 consecutive days in 2016, as shown in \rtab{DataCollection}. The data set is immensely huge, e.g., in Beijing it captures the trajectory of 31.8 million devices, which accounts for $\sim$50\% of the city's population. 
The spatial precision of each record is kept within 100 meters, by using the records collected by GPS and Wi-Fi. 

\begin{table}[t]
\centering
\small
\caption{The metadata of each urban trajectory record.}
\vspace{-0.1 in}
\label{tab:DataFormat}
\begin{tabular}{|l|l|l|}
\hline
Field                            & Description   & Sample   \\ \hline \hline
Time      & Timestamp of record & 18:02:41/07/12/2016   \\ \hline
Lon.      & Longitude of location & 116.523625            \\ \hline
Lat.      & Latitude of location & 39.792935             \\ \hline
Mid       & Unique device ID & 1370021020431    \\ \hline
%
\end{tabular}
\vspace{-0.1 in}
\end{table}

\begin{table}[t]
\centering
\small
\caption{Three trajectory data sets used in this work.}
\vspace{-0.1 in}
\label{tab:DataCollection}
\begin{tabular}{|l|l|l|l|l|}
\hline
City        & \#Device      & \#Record       & Size      & Length                      \\ \hline \hline
Beijing     & 31849742    & 8407648917     & 738.1G    & 90 days                \\ \hline
Tianjin     & 8011128     & 2858575880     & 206.8G    & 90 days                      \\ \hline
Tangshan    & 2786668     & 920364499      & 64.8G     & 90 days                      \\ \hline
\end{tabular}
\vspace{-0.1 in}
\end{table}


\bsubsec{The Long-Tailed Sparsity Pattern}{TrajectoryPattern}

We study the sampling statistics of the trajectory data. The time intervals between consecutive records inside the trajectory are averaged to $\sim$2.5 hours in all the three data sets. With these extremely sparse trajectories, it seems impossible to infer their continuous mobility. As a reference, most journeys in a city elapse no longer than two hours, during which less than one record is reported on average.

\begin{figure}[t]
\centering
\subfigure[]{\includegraphics[width=1.6 in]{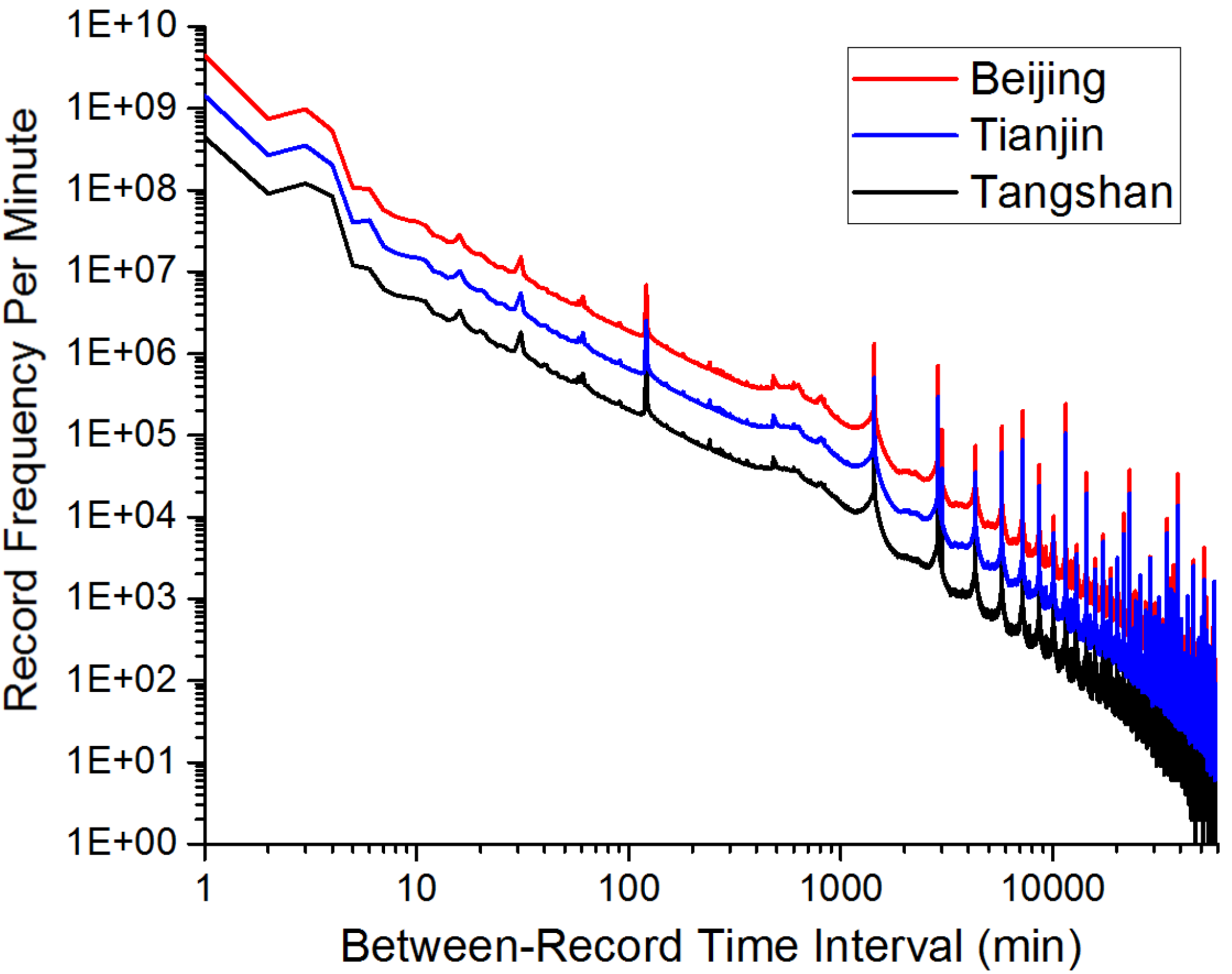}}
\subfigure[]{\includegraphics[width=1.44 in]{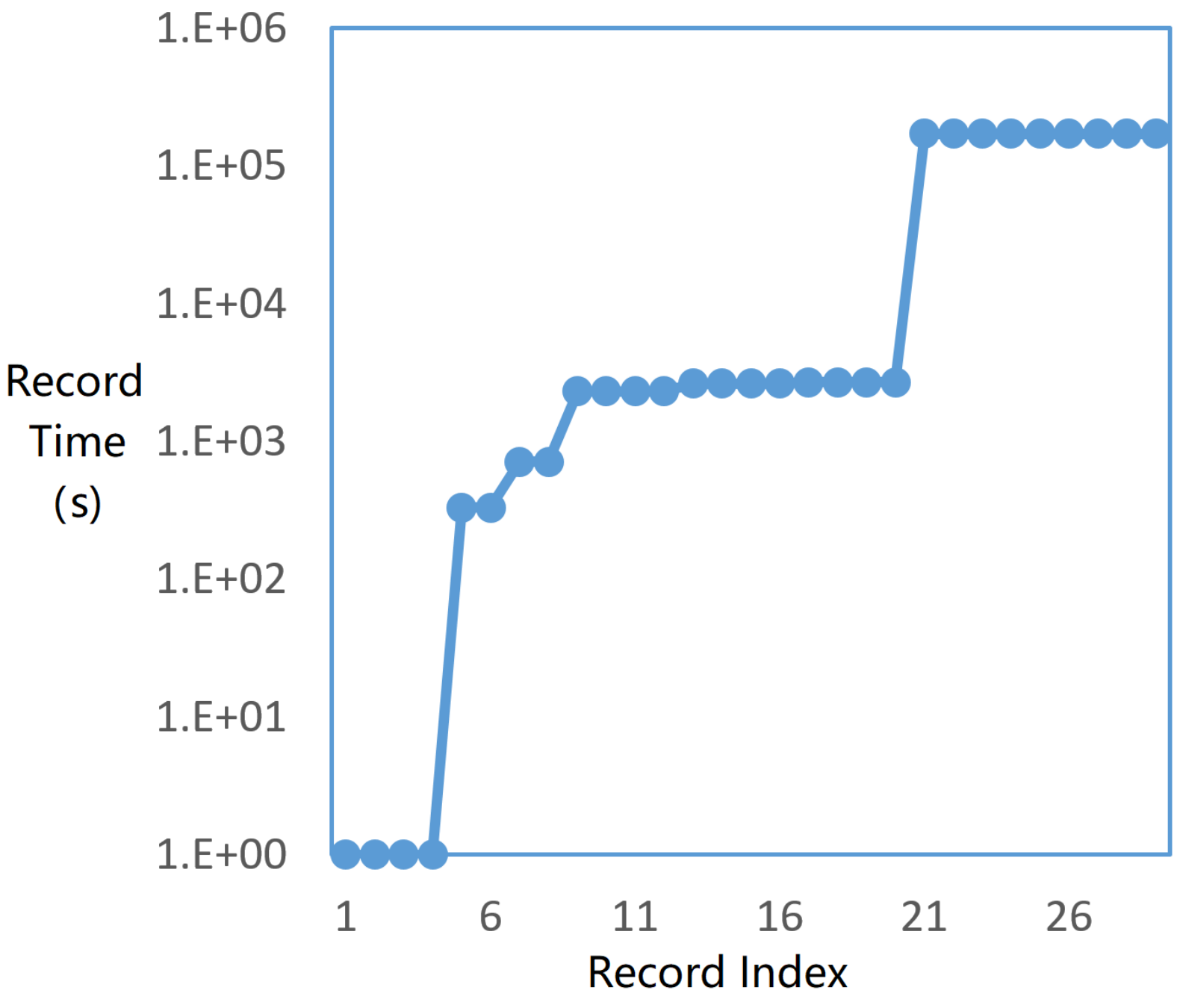}}
\vspace{-0.15 in}
\caption{The long-tailed sparsity pattern: (a) the distribution of between-record time intervals; (b) an example trajectory.}
\vspace{-0.22 in}
\label{fig:SampleTrajectory}
\end{figure}

Taking a closer look, we identify that the sampling pattern in our data set, though sparse, is highly skewed. \rfig{SampleTrajectory}(a) depicts the distributions of the between-record time intervals, which follow power-law like decays in the log-log scale. We call this pattern the long-tailed sparsity: most intervals are very short while there are also quite a few extremely long intervals that contribute to the large average. Take the Beijing data set as an example, 88.9\% intervals are smaller than 30 minutes (a typical $\Delta{T}$ for \rdef{Continuous-Segment}). At the trajectory level, it is observed that most trajectories are composed of multiple densely sampled segments that are far apart from each other over time. An example trajectory is depicted in \rfig{SampleTrajectory}(b).

To capture the long-tailed sparsity pattern, we define two metrics on each trajectory. These metrics are shown later to correlate with the capability for the mobility inference.

\bdefin{Sparsity}\textsc{Sparsity metrics of a trajectory} -- for any trajectory $\Gamma$ observed at $t \in \{t_1,\cdots,t_L\}$:

(a) \textbf{global sparsity} is the average time interval between the consecutive records of $\Gamma$: $\xi(\Gamma) = E(t_{i+1}-t_{i}), \forall i \in [1,L-1]$;



(b) \textbf{local coverage} is the ratio of the records within the dense segment: $\rho(\Gamma) =\frac{L-I(t_{i}-t_{i-1}> \Delta{T}, t_{i+1}-t_{i}>\Delta{T})}{L}$, $\forall i \in [2,L-1]$.


\edefin
Here $E(\cdot)$ and $I(\cdot)$ denote the mean function and the size of a set. Note that the global sparsity is independent of the temporal parameter $\Delta{T}$, while the local coverage is related to $\Delta{T}$.

\begin{figure}[t]
\centering
\subfigure[]{\includegraphics[height=1.35 in]{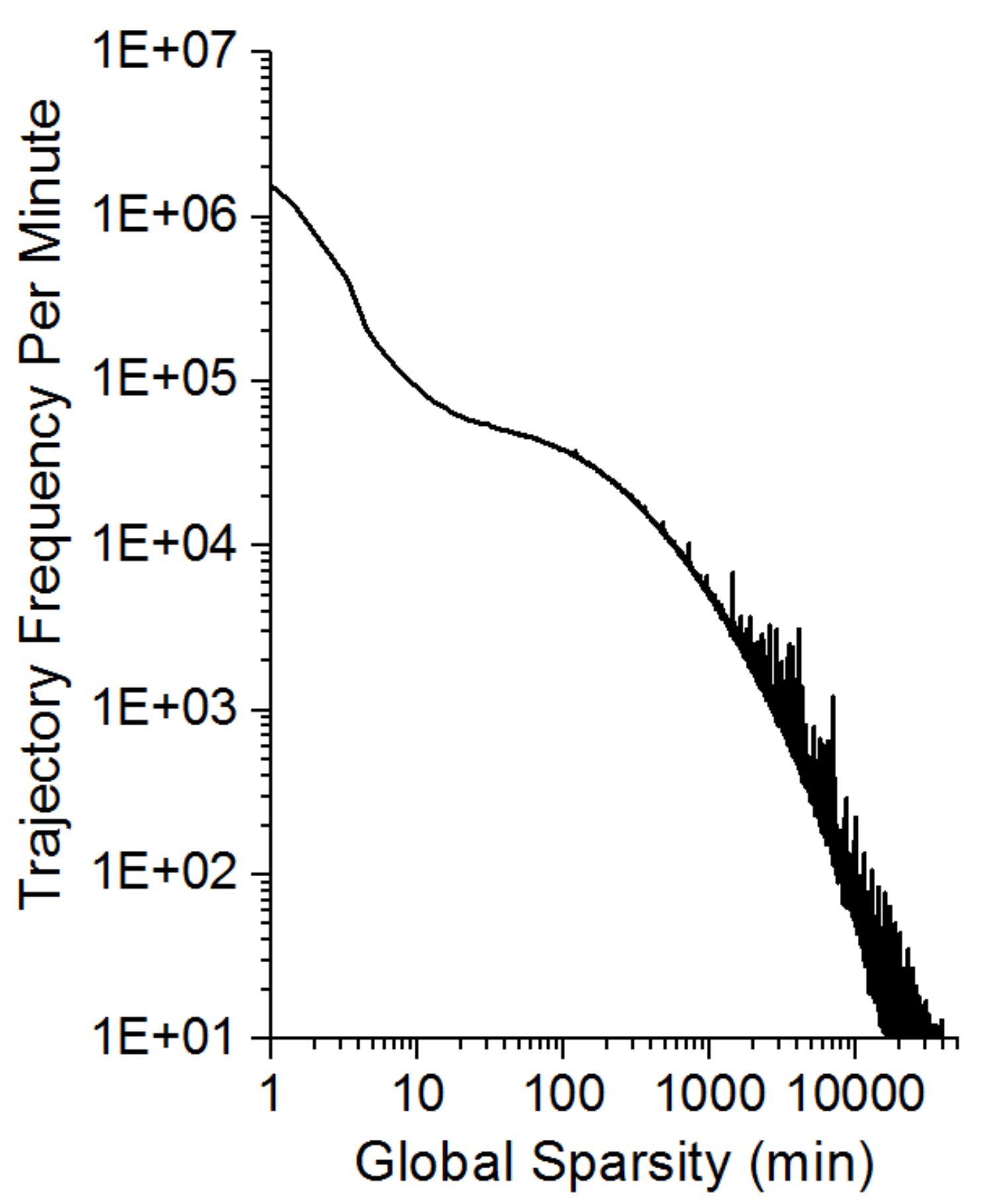}}
\hspace{-0.05 in}
\subfigure[]{\includegraphics[height=1.35 in]{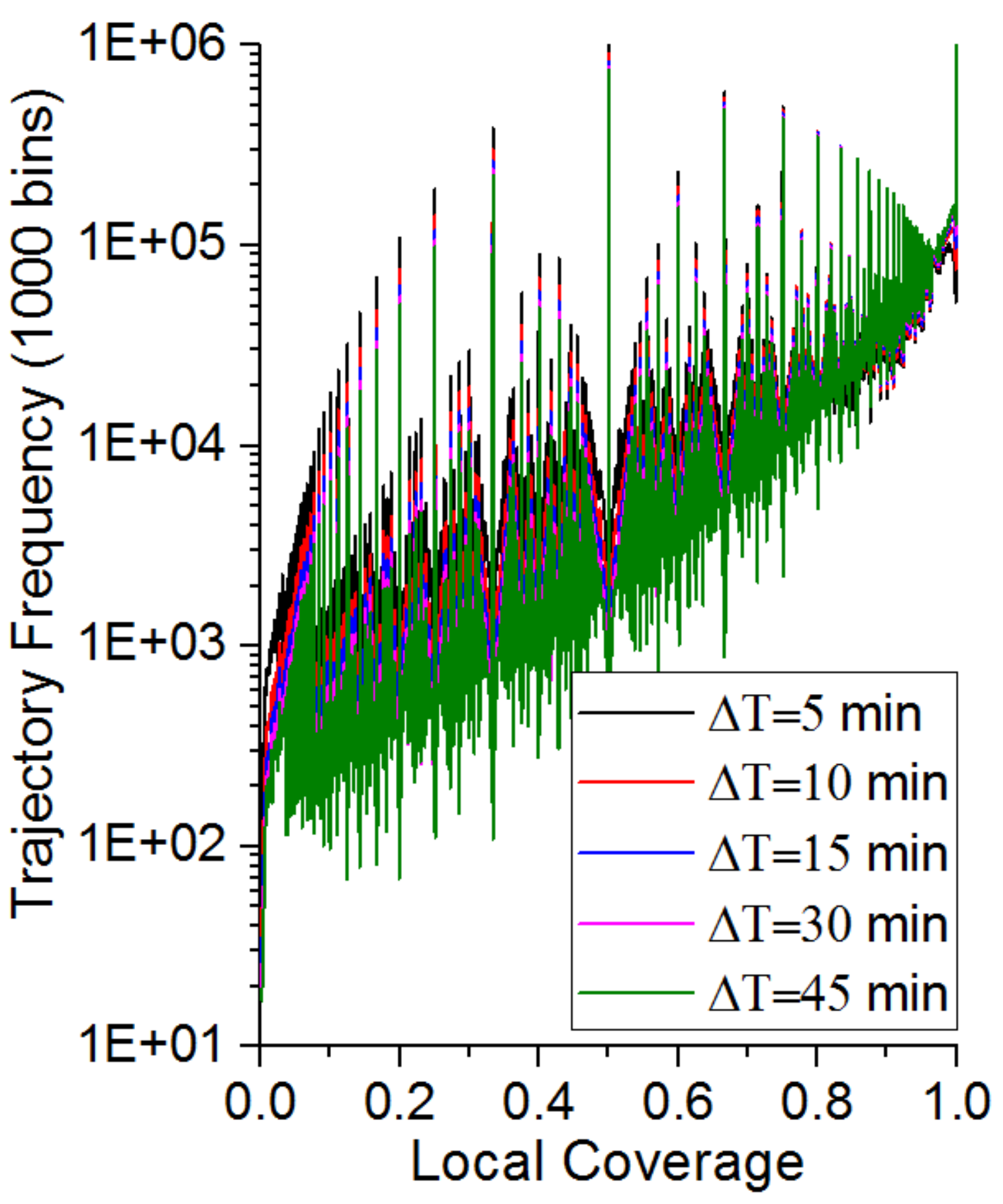}}
\hspace{-0.03 in}
\subfigure[]{\includegraphics[height=1.35 in]{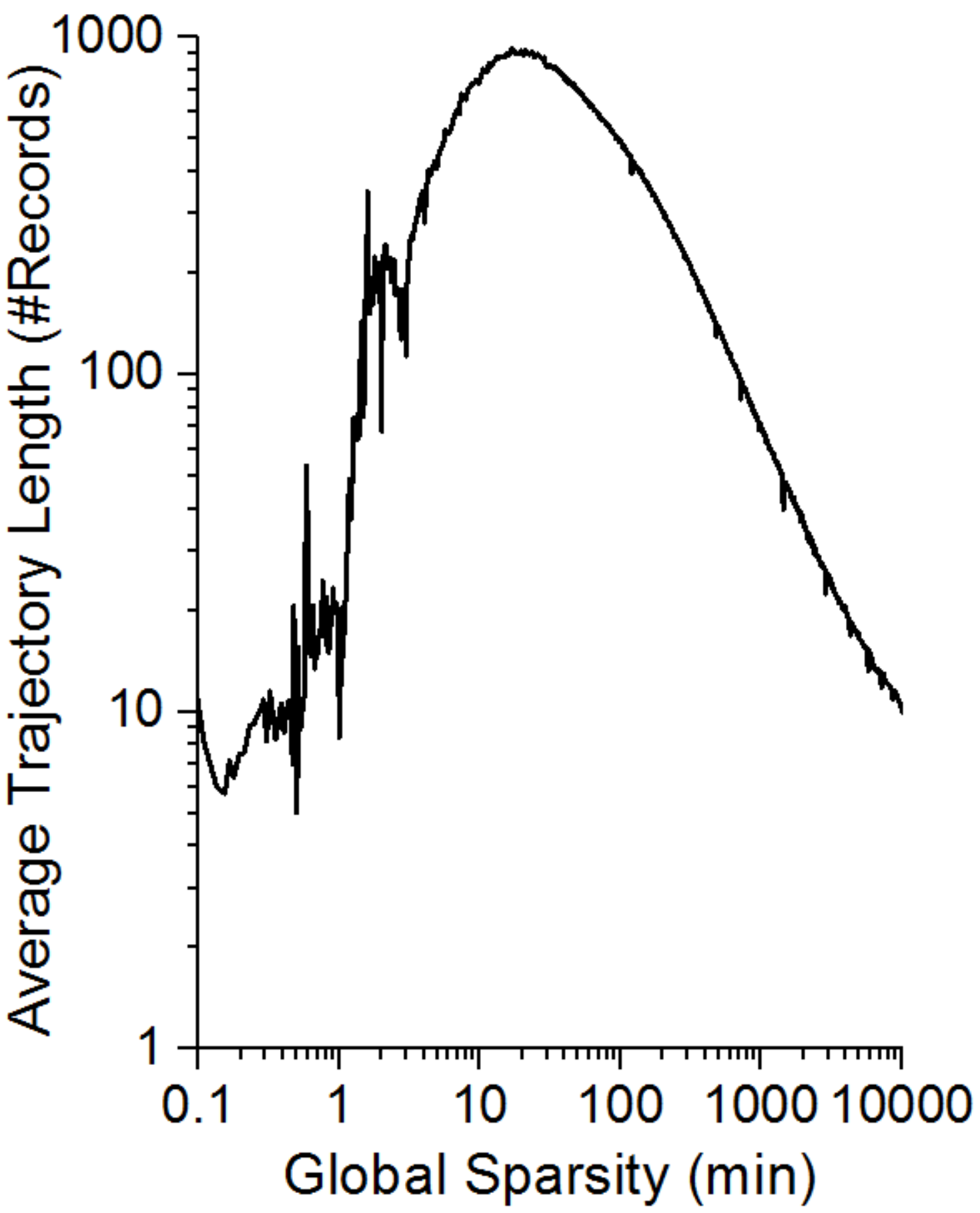}}
\vspace{-0.1 in}
\caption{The distribution of the sparsity metrics in the data of Beijing: (a) global sparsity; (b) local coverage; (c) average trajectory length by global sparsity.}
\vspace{-0.15 in}
\label{fig:TrajectorySparsity}
\end{figure}

We compute the sparsity metrics in the trajectory data of Beijing under five parameter settings ($\Delta{T}=5, 10, 15, 30, 45$ minutes). The distribution of the metrics are depicted in \rfig{TrajectorySparsity}. The global sparsity in \rfig{TrajectorySparsity}(a) follows a power-law like decay similar to the distribution of the between-record intervals in \rfig{SampleTrajectory}(a). 
The distribution of the local coverage in \rfig{TrajectorySparsity}(b) shows an exponentially increasing pattern that most of the trajectories have a high local coverage (with an average of 0.897 at $\Delta{T}=30$ minutes). This demonstrates that most of the records in the long-tailed sparse trajectory are in the densely sampled segment of the trajectory. As shown in \rfig{TrajectorySparsity}(c), the trajectories with extremely low and high sparsity tend to be smaller in length, i.e., the densely sampled short snippets or a few long-distance samples of a trajectory. The trajectories in both cases are therefore exempted from the subsequent analysis.

Note that the long-tailed sparsity pattern is also found in other data sets and application domains. For example, Gonzalez et al. studied the mobile phone user's trajectory data where the location of the user is reported upon each phone call or text message \cite{gonzalez2008understanding}. The time intervals between consecutive records follow a long-tailed power-law decay. In a recent work, Chen et al. analyzed the sparsely sampled geo-tagged social media data \cite{ChenSparse16}. The distributions of the time interval and distance between records follow power-law decays within the space and time scale of a single trip (1 day, 500km).

\bsec{Single Trajectory Inference}{Labeling}

\begin{algorithm}[t]
 \SetEndCharOfAlgoLine{}
\SetKwInOut{Input}{Input}\SetKwInOut{Output}{Output}
\newcommand\mycommfont[1]{\footnotesize{#1}}
\SetCommentSty{mycommfont}
\newcommand\mylnfont[3]{\footnotesize{#1}}
 \Input{$\Gamma=\bigcup_{i \in [1,L]}<t_i,\ell(t_i)>, t_1<\cdots<t_L$ (sparse trajectory), $\Delta{T}$, $\Delta{S}$ (space, time parameters)}
 \Output{$I_{S/T}(t_i), \forall i \in [1,L]$ (mobility of each record)}

\Begin{
 \tcc*[l]{$\Gamma$ into $M$ segments ($\gamma_j$) at every interval larger than $\Delta{T}$}
 $\{\gamma_j=\{<t_{j,k},\ell(t_{j,k})>\}_{k \in [1,L_j]}\}_{j=[1,M]} \leftarrow Divide(\Gamma, \Delta{T})$

 \For{$j \leftarrow [1,M]$,~$head \leftarrow 1$}{
     \tcc*[l]{detect all the stay segments on $\gamma_j$}

     \For{$cursor \leftarrow [2, L_j]$}{

         \tcc*[l]{iterate all the records backward from $cursor-1$}
         \For{$anchor \leftarrow [cursor-1,head]$}{

             \tcc*[l]{cut at the first escape outside the range of $\frac{\Delta{S}}{3}$}
             \If{$||\ell(t_{j,cursor})-\ell(t_{j,anchor})|| \geq \frac{\Delta{S}}{3}$}{

                 \tcc*[l]{stay segment}
                 \If{$t_{j,cursor-1}-t_{j,head} \geq \Delta{T}$}{
                     \For{$k \leftarrow [head, cursor-1]$}{
                         $I_{S/T}(t_{j,k}) \leftarrow$ S
                     }
                 }
                 $head \leftarrow anchor+1$,~\Break
             }
         }
     }
     \tcc*[l]{detect all the travel records on $\gamma_j$}
     \For{$cursor \leftarrow [2, L_j-1]$ \&\& $I_{S/T}(t_{j,cursor}) \ne$ S}{
         \tcc*[l]{find the first left record outside the range of $\Delta{S}$}
         \For{$l \leftarrow [cursor-1, 1]$}{
             \If{$||\ell(t_{j,cursor})-\ell(t_{j,l})|| \geq \Delta{S}$}{
                 $left \leftarrow l$,~\Break
             }
         }
         \tcc*[l]{find the first right record outside the range of $\Delta{S}$}
         \For{$r \leftarrow [cursor+1, L_j]$}{
             \If{$||\ell(t_{j,cursor})-\ell(t_{j,r})|| \geq \Delta{S}$}{
                 $right \leftarrow r$,~\Break
             }
         }
         \If{$t_{j,right}-t_{j,left} \leq \Delta{T}$}{
             $I_{S/T}(t_{j,cursor}) \leftarrow$ T
         }
     }
 }
 \Return $I_{S/T}(t_i), i=[1,L]$
}
\caption{SDS on long-tailed sparse trajectories.}
\label{alg:SDS}
\vspace{-0.05 in}
\end{algorithm}




We propose the mobility inference algorithm on the single trajectory. The main idea is to leverage the long-tailed sparsity pattern discovered in our trajectory data (\rsubsec{TrajectoryPattern}). Though the average record interval in a trajectory is too large to apply \rthe{Continuous-Discrete}, each trajectory can be decomposed into multiple densely sampled segments, on which the continuous mobility can be confidently inferred.


\bdefin{Dense-Segment}\textsc{Dense stay segment} -- a discrete segment $\gamma$ of the trajectory $\Gamma$ defined in the time series $\omega = \{t_p,\cdots,t_q\}\;(1 \leq p < q \leq L)$ is a \textbf{dense stay segment} of $\Gamma$ if:

(a) $\gamma$ is a stay segment of $\Gamma$ by \rdef{Discrete-Segment}(a);

(b) any consecutive time interval of $\gamma$ is small enough: $\forall p \leq i < q,\;t_{i+1}-t_i \leq \Delta{T}$. $\Delta{T}$ is the parameter used in \rdef{Discrete-Segment}(a).
\edefin

\bprop{ContinuousAssumption}\textsc{Continuous stay assumption} -- Consider a dense stay segment $\gamma$ detected from the long-tailed sparse trajectory, which is defined in the time series $\omega = \{t_p,\cdots,t_q\}\;(1 \leq p < q \leq L)$. For any unobserved time point $t \in (t_i,t_{i+1}), \forall p \leq i < q$, we hypothesize that $||\ell(t)-\ell(t_i)||<\Delta{S}$ and $||\ell(t)-\ell(t_{i+1})||<\Delta{S}$.
\eprop


\rprop{ContinuousAssumption} states that if a user is observed frequently in a region of diameter $\Delta{S}$, any intermediate location between observations is also within a similarly constrained region. We empirically validate this observation by the experiment in \rsuba{Proposition} on our trajectory data set. The probability of violating the observation is below $10^{-5}$ in most cases. When \rprop{ContinuousAssumption} holds, we can develop two theorems that characterize the continuous mobility of stay and travel on long-tailed sparse trajectories.


\bthe{Continuous-Dense}\textsc{Continuous mobility of dense stay segments} -- In the long-tailed sparse trajectory $\Gamma$:

(a) any dense stay segment $\gamma$ satisfying \rdef{Dense-Segment} under the parameters of $\Delta{S}/3$ and $\Delta{T}$ is also the stay segment by \rdef{Continuous-Segment}(a) in the continuous model under the parameters of $\Delta{S}$ and $\Delta{T}$;

(b) the continuous mobility of any discrete segment $\gamma$ in the time period $\tau \in [t_p,t_q]$ can be inferred as stay by \rdef{Continuous-Segment}(a) under the parameters of $\Delta{S}$ and $\Delta{T}$ only if $\gamma$ defined in $\omega = \{t_p,\cdots,t_q\}$ is the dense stay segment under the same parameters.
\ethe


\bthe{Continuous-Travel}\textsc{Continuous mobility of travel records} -- Consider a discrete trajectory $\Gamma$ defined in the time series $\omega = \{t_1,\cdots,t_L\}$:

(a) any record at time $t_i\;(1<i<L)$ is in the travel segment by the continuous model of \rdef{Continuous-Segment}(b) under the parameters of $\Delta{S}$ and $\Delta{T}$ if only there exist $1 \leq p<i<q\leq L$ that: 1) $||\ell(t_i)-\ell(t_{p})|| \geq \Delta{S}$; 2) $||\ell(t_i)-\ell(t_{q})|| \geq \Delta{S}$; 3) $t_q-t_p \leq \Delta{T}$;

(b) any record at time $t_i$ can be inferred as in the travel segment by \rdef{Continuous-Segment}(b) under the parameters of $\Delta{S}$ and $\Delta{T}$ only if there exist $1 \leq p<i<q\leq L$ that: 1) $||\ell(t_i)-\ell(t_{p})|| \geq \Delta{S}/2$; 2) $||\ell(t_i)-\ell(t_{q})|| \geq \Delta{S}/2$; 3) $t_q-t_p \leq \Delta{T}$.
\ethe


The proofs are given in \rsuba{Proof}. By \rthe{Continuous-Dense} and \rthe{Continuous-Travel}, we design a new algorithm to infer the continuous mobility of a single long-tailed sparse trajectory, called Slice \& Doubly Sliding (SDS). As shown in \ralg{SDS}, the algorithm first slices the trajectory into multiple dense segments at all the intervals larger than $\Delta{T}$ (L2). On each dense segment $\gamma$, the stay/travel segments are detected respectively (L3$\sim$10, L11$\sim$19). In particular, the stay detection checks all the segments of $\gamma$ with the condition in \rdef{Dense-Segment} under the parameters of $\Delta{S}/3$ and $\Delta{T}$ by \rthe{Continuous-Dense}. To avoid the worst-case $O(L^4)$ complexity, we introduce a doubly sliding window data structure which keeps track of the currently checked segment. The key of the algorithm lies in that, when one pair of records no closer than $\Delta{S}/3$ are found (L6), all the segments containing this pair of records will be pruned early in the detection and the sliding window will advance aggressively (L10). The travel detection follows \rthe{Continuous-Travel}. The average-case complexity of SDS is $O(L \cdot W)$ where $W$ is the average number of records in a maximal stay segment.

According to \rthe{Continuous-Dense}(a) and \rthe{Continuous-Travel}(a), the SDS algorithm guarantees a 100\% precision in the mobility inference of both stay and travel. By \rthe{Continuous-Dense}(b) and \rthe{Continuous-Travel}(b), the lower bounds of the recalls in detecting the stay and travel are $\frac{SDS(\Gamma, S, \Delta{S}/3, \Delta{T})}{SDS(\Gamma, S, \Delta{S}, \Delta{T})}$ and $\frac{SDS(\Gamma, T, \Delta{S}, \Delta{T})}{SDS(\Gamma, T, \Delta{S}/2, \Delta{T})}$ respectively, where $SDS(\Gamma, S/T, \Delta{S}, \Delta{T})$ are the number of stay and travel records detected by the SDS algorithm from $\Gamma$ under the parameters of $\Delta{S}$ and $\Delta{T}$. Note that the recall is defined on all the stay and travel records that can be detected given the single sparse trajectory, not on the continuous mobility of records given the full trajectory information. Another advantage of the SDS algorithm lies in that it also works for dense trajectories, each of which is treated as one densely sampled segment.

\begin{figure}[t]
\centering
\subfigure[]{\includegraphics[width=1.6 in]{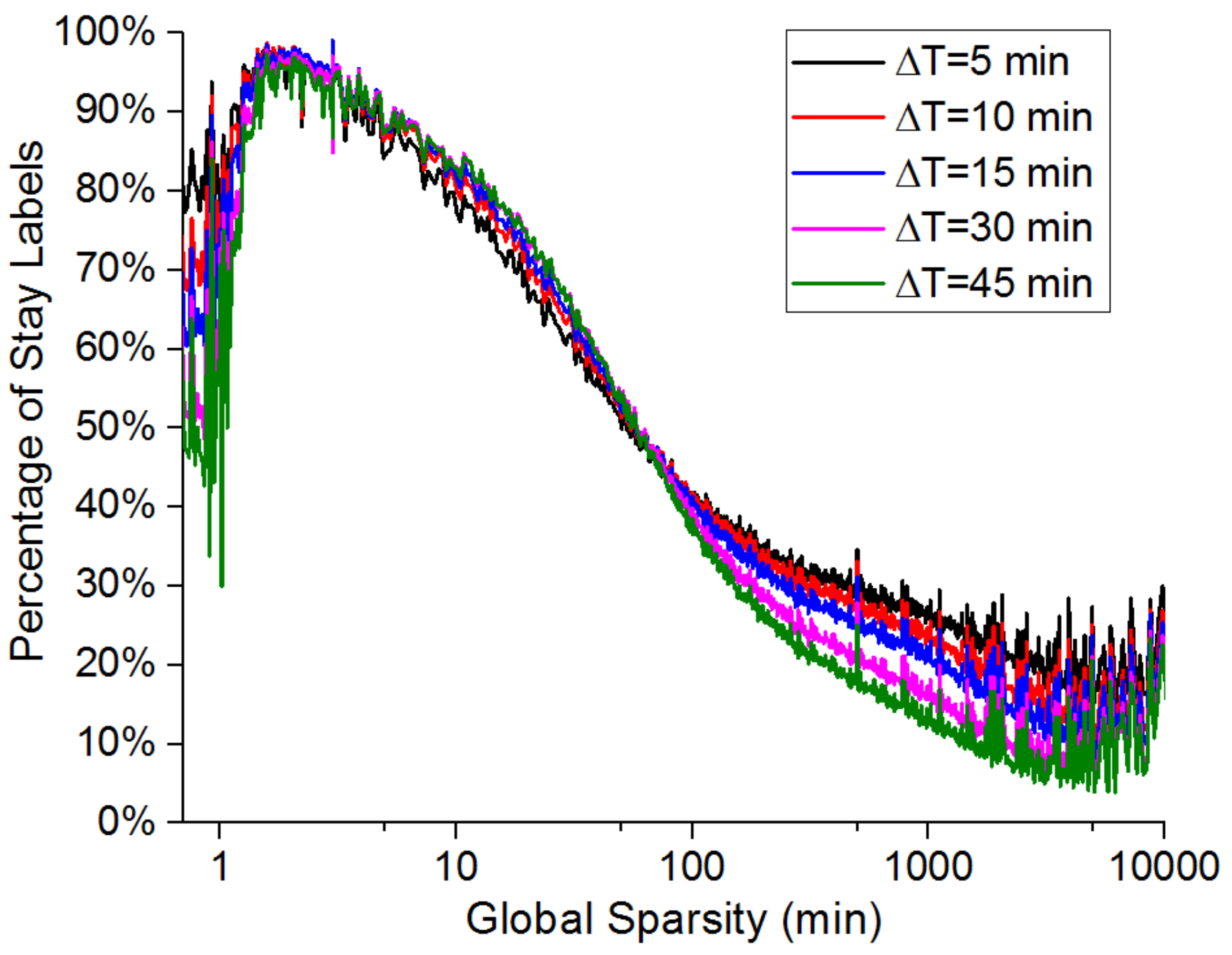}}
\subfigure[]{\includegraphics[width=1.6 in]{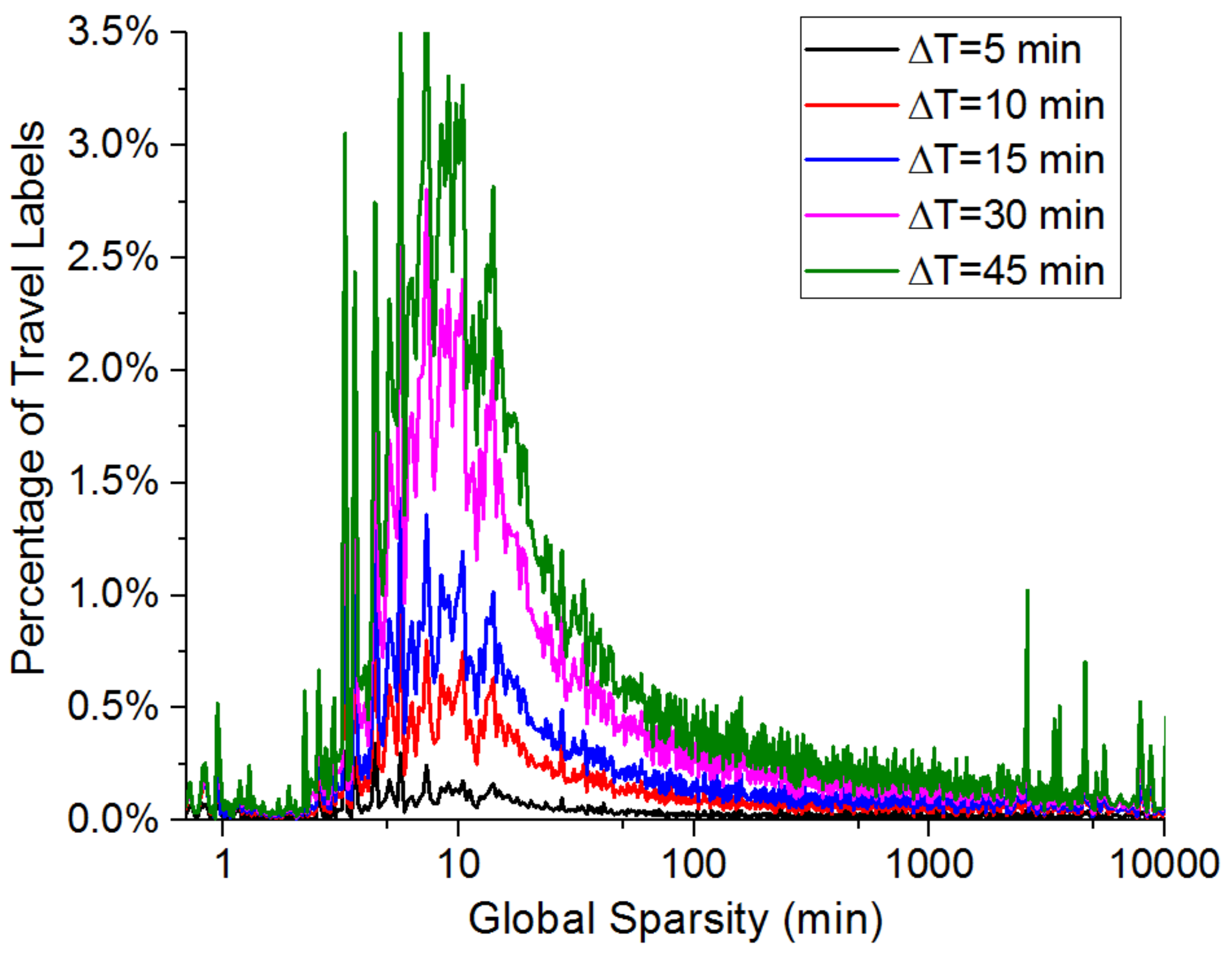}}\\
\vspace{-0.15 in}
\caption{The percentage of stay/travel on the trajectory with different global sparsity values and $\Delta{T}$ ($\Delta{S}$ fixed to 800 m).}
\vspace{-0.2 in}
\label{fig:TrajectoryLabeling}
\end{figure}

We applied the SDS algorithm to our data set in Beijing. 47.0\%$\sim$50.2\% and 0.044\%$\sim$0.83\% records are detected as stay and travel, depending on the parameters of $\Delta{S}$ and $\Delta{T}$. \rfig{TrajectoryLabeling} shows the average stay/travel percentages by the global sparsity of a trajectory. All the curves are bell-shaped with only one peak: the highest ratio of stay is found at the global sparsity around 1.6 min (97.3\%$\sim$98.5\%, \rfig{TrajectoryLabeling}(a)); the highest ratio of travel is found at the global sparsity from 5 min to 10 min, which increases with $\Delta{T}$ (0.34\%$\sim$3.8\%, \rfig{TrajectoryLabeling}(b)). 
Before the peak of stay, the trajectory is mostly composed of less than 10 records (\rfig{TrajectorySparsity}(c)), with a time period shorter than $\Delta{T}$ and can not be inferred as stay. After the peaks of stay and travel, the ratio of detected records drops due to the increased sparsity of the trajectories. This validates \rthe{Continuous-Discrete} that sparser trajectories are harder for the continuous mobility inference.

By the empirical result, the parameters of $\Delta{T} = 30 min$ and $\Delta{S} = 800 m$ are chosen and used throughout this work. The details are explained in \rsuba{Proposition}.

\bsec{Multiple Trajectory Inference}{Model}



The SDS algorithm correctly infers the continuous mobility of 47.8\%$\sim$50.3\% records in each trajectory of our data set. For most of the other records, the inference is not feasible given the single trajectory only, as proved in \rthe{Continuous-Dense}(b) and \rthe{Continuous-Travel}(b). We propose to employ the deep learning based inference model which trains over multiple trajectories to learn the spatiotemporal regularity in human mobility beyond the definitive rules in the SDS algorithm.

The recurrent neural network (RNN), or specifically LSTM \cite{LSTM}, is a classic model to analyze the sequence data. Nevertheless, there are two issues to directly apply LSTM in our scenario: 1) the single LSTM network maps an input sequence to an output sequence of the same length. By default, only the list of records labeled by the SDS algorithm can be used as the input; 2) the trajectories of all the users are first spliced in tandem and then cut into fix-size slices for training. The local trajectory context, instead of the per-user context, is used for inference in the single LSTM network.

\bsubsec{Trajectory Encoder and Mobility Decoder}{EncoderDecoder}

\begin{figure}[t]
\centering
\includegraphics[width=2.5 in]{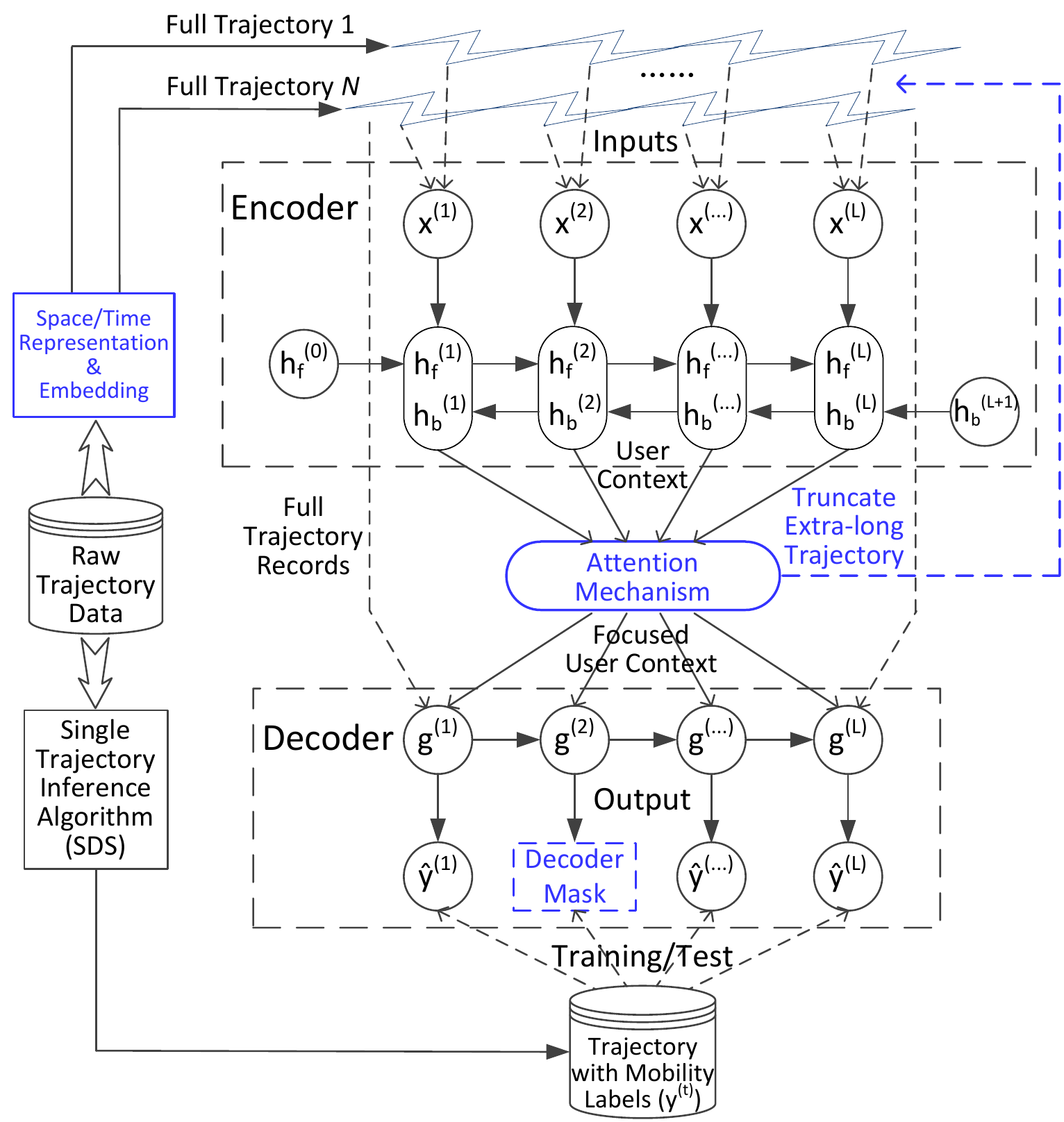}
\vspace{-0.1 in}
\caption{The encoder-decoder sequence-to-sequence learning architecture for multiple trajectory inference. Blue outlines indicate the optimized design for our problem.}
\vspace{-0.15 in}
\label{fig:Model}
\end{figure}

We propose an encoder-decoder architecture for the sequence-to-sequence learning \cite{cho2014learning}\cite{sutskever2014sequence} to overcome the limitations of the single LSTM network. As shown in \rfig{Model}, the encoder summarizes the full trajectory of each user into the per-user context with a bidirectional LSTM network (the upper part of the figure). Based on the user context, the decoder takes records from the same trajectory and sequentially infers their mobility by another unidirectional LSTM network (the lower part of the figure). 


Formally, for a trajectory $\Gamma$ of length $L$, the encoder at the time step of $t=[1,L]$ are defined by
\beq{Encoder-1}
\textbf{h}_{f}^{(t)} = tanh(\textbf{b}_f+\textbf{W}_f\textbf{h}_{f}^{(t-1)}+\textbf{U}_f\textbf{x}^{(t)})\nonumber
\eeq
\beq{Encoder-2}
\textbf{h}_{b}^{(t)} = tanh(\textbf{b}_b+\textbf{W}_b\textbf{h}_{b}^{(t+1)}+\textbf{U}_b\textbf{x}^{(t)})
\eeq
where $\textbf{h}_{f}^{(t)}$ and $\textbf{h}_{b}^{(t)}$ denote the hidden states of the forward/backward LSTM at the time step $t$ respectively, $\textbf{x}^{(t)}$ is the input record of the trajectory at $t$ using the space/time representation (\rsubsec{Embedding}).

In the decoder side, the operation is defined by
\beq{Decoder-1}
\textbf{g}^{(t)} = tanh(\textbf{b}_d+\textbf{W}_d\textbf{g}^{(t-1)}+\textbf{U}_d\textbf{x}^{(t)}+\textbf{V}_d\textbf{C}^{(t)}),~\textbf{g}^{(0)}=\textbf{0}\nonumber
\eeq
\beq{Decoder-2}
\hat{\textbf{y}}^{(t)} = softmax(\textbf{c}+\textbf{W}_o\textbf{g}^{(t)})
\eeq
where $\textbf{g}^{(t)}$ denotes the hidden state of the decoder LSTM, $\textbf{C}^{(t)}$ is the focused context at the time step $t$ by the attention mechanism (\rsubsec{Attention}), $\hat{\textbf{y}}^{(t)}$ is the predicted mobility distribution at $t$.

\bsubsec{Space/Time Representation and Embedding}{Embedding}

In the raw trajectory data, both space and time information are recorded in high resolutions, i.e., millionths of a second (timestamp) or a degree (longitude/latitude). To cope with the neural network input, we select the appropriate spatial/temporal features, discretize their values, and compute the vector embeddings to represent them. The embeddings are updated online during the training process. 


In space, we divide the territory of each city into grids of $1\textperthousand$ degree latitude/longitude. The location of each record is converted to the latitude and longitude indices of the grid it belongs to. Each grid index is represented by a vector of length $E$. In time, we divide the timestamp of each record into two indices and embed them separately. The first is the absolute hour index of the timestamp. For example, the timestamp of 12:50AM on Monday has an index of $12+24 \cdot 1=36$. The hour index indicates the time of day and the day of the week upon the observation, which can be related to the mobility of the record. The second is the relative minute index defined as the elapsed time in minutes from the start of the current segment in the trajectory. Here the segments of a trajectory is computed by L2 of \ralg{SDS}. Both the hour and the minute indices have a length of $E$ in the embedding. Finally, the space and time embeddings are concatenated into a vector of length $4E$ as the input to the encoder-decoder architecture, i.e., $\textbf{x}^{(t)}=[Grid(\ell(t)); Traj\_Hour(t); Seg\_Minute(t)]$. The embedding for each grid/time index is randomized upon the initialization. We use $E=100$ by default.

\bsubsec{Attention Mechanism}{Attention}


In the encoder side of our architecture, the input trajectory on average can be too long to be summarized as a fixed length context vector. We introduce the attention mechanism \cite{luong2015effective} where the encoder produces an array of context vectors associated with each record and saves it in a short-term memory. The decoder builds a neural network to dynamically attend the memory and compute the context vector which is used in the inference of each record. 

The focused context $\textbf{C}^{(t)}$ at the time step $t$ is computed by
\beq{Attention-1}
score(\textbf{g}^{(t)}, \textbf{h}^{(s)})= \textbf{g}^{(t)\top}\textbf{W}_a\textbf{h}^{(s)},~\textbf{h}^{(s)} = [\textbf{h}_{f}^{(s)};\textbf{h}_{b}^{(s)}]\nonumber
\eeq
\beq{Attention-2}
\textbf{C}^{(t)}=\sum_{s=[1,L]}{\textbf{a}_{ts}\textbf{h}^{(s)}},~\textbf{a}_{ts} = \frac{exp(score(\textbf{g}^{(t)}, \textbf{h}^{(s)}))}{\sum_{s'=[1,L]}{exp(score(\textbf{g}^{(t)}, \textbf{h}^{(s)}))}}
\eeq
where $\textbf{A}=\{\textbf{a}_{ts}\}_{t,s=[1,L]}$ is the attention matrix.

In a typical attention-based sequence-to-sequence learning architecture, e.g., neural machine translation \cite{bahdanau2014neural}, the input sequence is a sentence of 100 words at maximum. While in our problem, the user trajectory could be much longer, up to 5000 records. Therefore, we further optimize the learning procedure by truncating each long trajectory into multiple fix-length sub-trajectories. A local context vector is trained on each sub-trajectory, as proposed by Luong et al. \cite{luong2015effective}. The record in the decoder side will only attend to the records belonging to the same sub-trajectory in the encoder, thus resolves the issue with extra-long trajectories.

\bsubsec{Train/Test with Decoder Masks}{Mask}


In the decoder side, we only have the mobility label (stay or travel) for a subset of records on each trajectory. If we only include these labeled records in the training, the test performance might downgrade because of the loss of consistent context in the decoder side, as empirically shown by the performance of the baseline LSTM in \rsubsec{Result}. In our improved design, we still feed the full trajectory to the decoder side in the training, but mask out the losses for the records without labels. This allows the decoder LSTM network consistently capture the dynamics of the entire trajectory and be trained with supervisions if available.

The final objective function for training is
\beq{DecoderMask}
\digamma=-\sum_{t \in S_{label}}{\log{p(\textbf{y}^{(t)}|\hat{\textbf{y}}^{(t)})}}
\eeq
where $\textbf{y}^{(t)}$ denotes the mobility label at the time step $t$, $S_{label}$ is the set of record indices labeled by the SDS algorithm.

\bsec{Evaluation}{Eva}

\bsubsec{Experiment Setup}{Prep}

\textbf{Data.} We evaluate the SDS algorithm and encoder-decoder model on three types of data extracted from the raw data in \rsubsec{Source}.


\begin{itemize}[leftmargin=.1in]

\item \textit{\textbf{Full}} data is the set of randomly selected trajectories. We apply the SDS algorithm to create the stay/travel labels on each trajectory. Because the travel labels are rare ($<1\%$) and a large percentage of short trajectories have no travel label at all, we only select the trajectories with at least 10 travel labels. Note that this criterion does not lead to a biased selection for the mobility inference. The eligible trajectories have an average global sparsity mildly smaller than the average in the whole data set. We extract $2 \times 3$ groups of non-overlapping 10K, 40K, 100K trajectories for train and test respectively. They are called \texttt{FU-10K}, \texttt{FU-40K}, \texttt{FU-100K}. By default, the Beijing data is used. 
Only the labeled records on the full data can be evaluated.

\item \textit{\textbf{Re-sampled}} data is used to evaluate the inference performance on unlabeled records. Given a trajectory in the full data, we randomly keep each record with a probability (i.e., the re-sampling rate). The re-sampled training data is re-labeled by the SDS algorithm, normally generating a smaller percentage of labels than the full data. On the re-sampled test data, we re-use the labels generated in the full data. The series of data re-sampled from \texttt{FU-10K} is called \texttt{RE-10K}. The records with labels in the full data can be evaluated.

\item \textit{\textbf{Simulated}} data is used to evaluate the SDS algorithm itself, as no true label can be detected beyond the algorithm. The simulated data is generated using the timestamps in the full data and then re-sampled (described in \rsuba{Simulation}). The labels of all the records in the simulated data are known and can be evaluated.

\end{itemize}

\noindent \textbf{Method.} Seven mobility inference methods are compared.

\begin{itemize}[leftmargin=.1in]

\item \textit{\textbf{LSTM.}} The unidirectional LSTM cell is used as the baseline of the sequence deep learning method. Only labeled records are fed as input in both train and test.
\item \textit{\textbf{Voting.}} A spatiotemporal bin is combined by a spatial grid and an hour index defined in \rsubsec{Embedding}. All the labeled records in the training data are categorized into these bins and counted. The more frequent mobility type in each bin is used as the prediction for all the test records in this bin. For bins without a labeled record, a random prediction is computed.
\item \textit{\textbf{LG.}} To use traditional classifiers, we conduct the window-based feature extraction on each record. Within the dense segment generated by L2 of \ralg{SDS}, $W$ records are uniformly selected before and after the current record, forming a feature vector of length $(2W+1) \cdot 4$ where the four indices in \rsubsec{Embedding} are used to represent each record. We test through $W=1\sim10$ and find that $W=4$ achieves the best trade-off between performance and cost.
\item \textit{\textbf{DT.}} It applies the decision tree \cite{breiman1984classification} over the above features.
\item \textit{\textbf{NB.}} It applies the Gaussian Naive Bayes.
\item \textit{\textbf{L-SVM.}} It applies the linear SVM as the kernel SVM is not scalable to millions of records.
\item \textit{\textbf{HMM.}} The stay/travel is used as the hidden state, the spatiotemporal offset between consecutive records is used as the observation. The prediction is computed by the Viterbi algorithm \cite{viterbi1967error}. The method mimics the technique in the next location prediction using Markov chains \cite{gambs2012next}.
\end{itemize}

In each test, we measure the $P$recision and $R$ecall in predicting the $S$tay and tra$V$el labels separately, which are denoted as $SP$, $SR$, $VP$, $VR$, and the overall accuracy as $ACC$. As the percentage of labels is unbalanced (much fewer travels than stays), we also report $F1-ACC$, the harmonic mean of the F1 measures for stay and travel prediction.

\bsubsec{Quantitative Result}{Result}

\begin{figure}[t]
\vspace{-0.11 in}
\centering
\subfigure[]{\includegraphics[width=1.5 in]{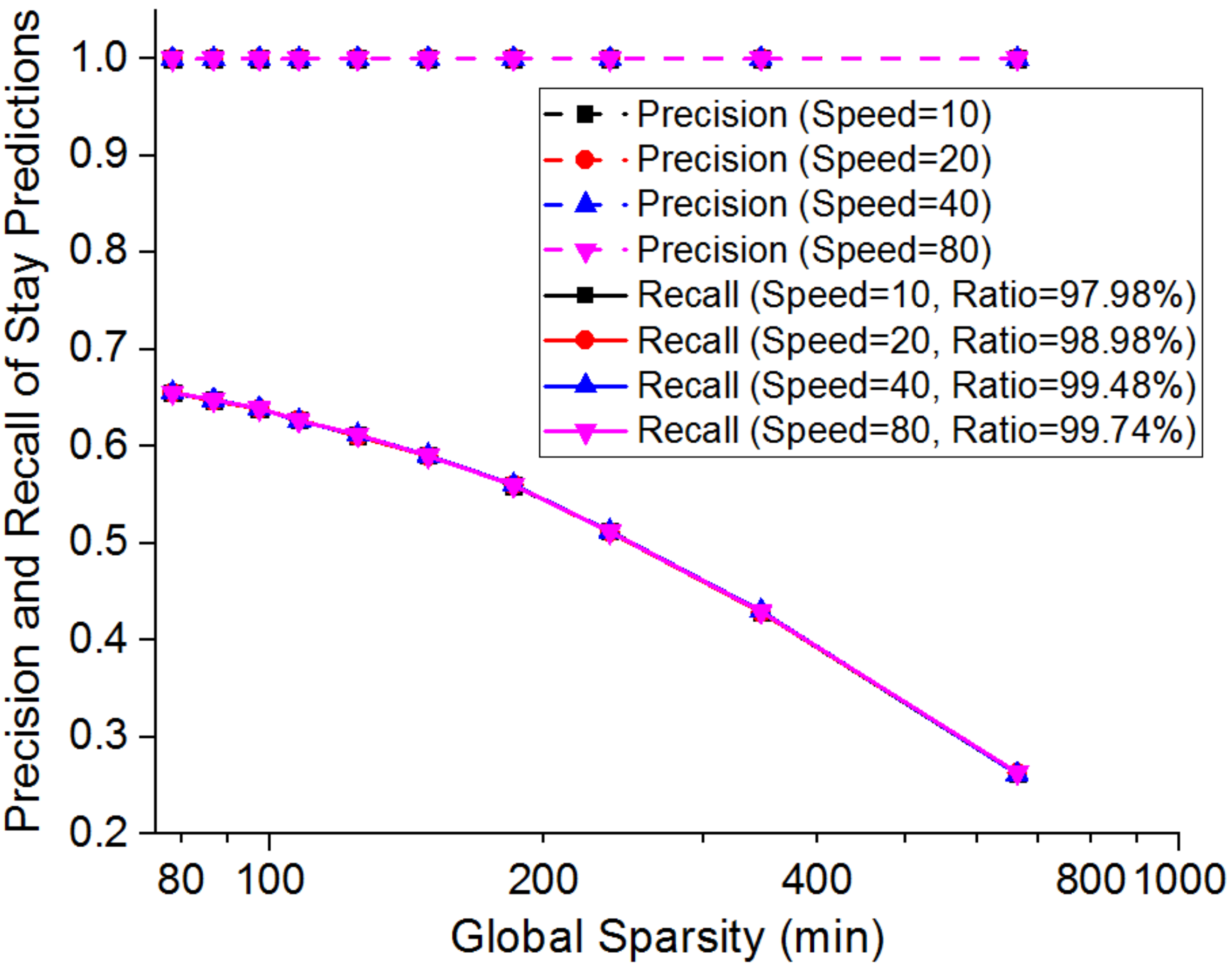}}
\subfigure[]{\includegraphics[width=1.5 in]{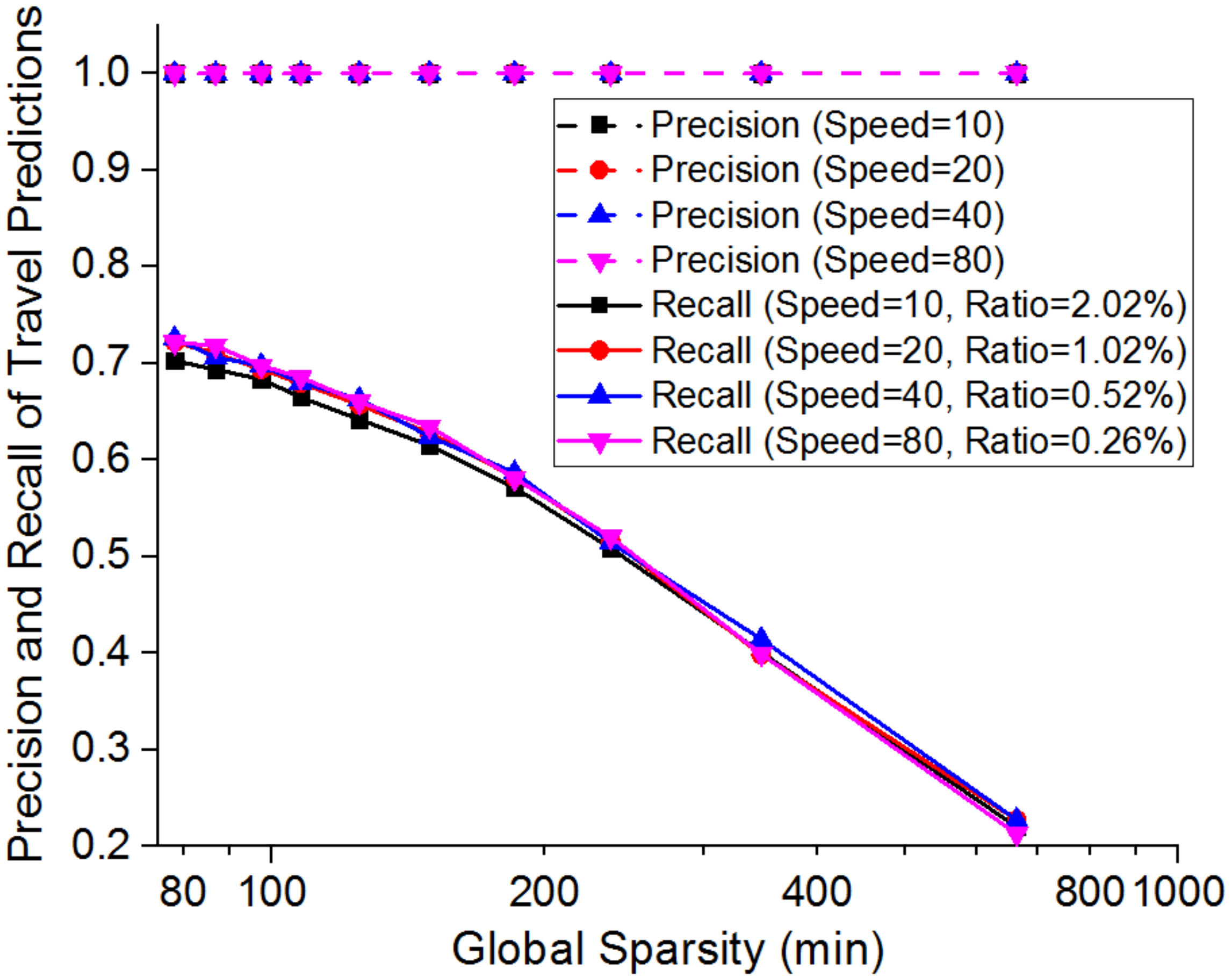}}
\vspace{-0.15 in}
\caption{SDS inference on simulated data: (a) stay; (b) travel. }
\vspace{-0.2 in}
\label{fig:SDSPerformance}
\end{figure}

\textbf{SDS algorithm.} We evaluate the SDS algorithm on the simulated data over \texttt{RE-10K}, with re-sampling rates from 1.0 to 0.1. As shown in \rfig{SDSPerformance}, the precision of both stay and travel predictions (dashed lines) is 100\%, regardless of the re-sampling rate and the speed used in the simulation. This validates the theoretical result in \rsec{Labeling}. As the re-sampling rate decreases, which leads to a linear increase in the global sparsity by \rdef{Sparsity}(a) (X axis), the recall drops in a rate slightly slower than the empirical result in \rfig{TrajectoryLabeling}.

\begin{table*}[t]
\small
\centering
\caption{Performance with different encoder-decoder design parameters on \texttt{FU-10K} data.}
\vspace{-0.1 in}
\label{tab:ModelParameter}
\begin{tabular}{p{.3cm}|p{1cm}|p{.35cm}p{.35cm}p{.35cm}|p{.35cm}p{.35cm}p{.35cm}|p{.35cm}p{.35cm}p{.35cm}|p{.35cm}p{.8cm}p{.8cm}|p{.35cm}p{.35cm}|p{.4cm}p{.8cm}}
\hline
\multicolumn{2}{c|}{\multirow{2}{*}{}}    & \multicolumn{3}{c|}{\# NN Layers} & \multicolumn{3}{c|}{Dropout Prob.} & \multicolumn{3}{c|}{Truncate Size} & \multicolumn{3}{c|}{Optimizer} & \multicolumn{2}{c|}{Attention} & \multicolumn{2}{c}{GRU}\\ \cline{3-18}
\multicolumn{2}{c|}{}                     & 1         & 2         & 3        & 0          & 0.2       & 0.5       & 100    & 200  & 400        & SGD    & adagrad  & adadelta         & with  & w/o & with att. & w/o  \\ \hline\hline
\multirow{2}{*}{Pr.} & Stay ($SP$)   & 0.97	& 0.95	& 0.96	& 0.97	& 0.96	& 0.85	& 0.97	& 0.97	& 0.87	& 0.97	& 0.85	& 0.95	& 0.97	& 0.95 & 0.95 & 0.95\\ 
                           & Travel ($VP$) & 0.82	& 0.78	& 0.78	& 0.82	& 0.80	& 0.00	& 0.88	& 0.82	& 0.61	& 0.82	& 0.00	& 0.79	& 0.82	& 0.79 & 0.82 & 0.80 \\ \hline
\multirow{2}{*}{Re.}    & Stay ($SR$)  & 0.97	& 0.96	& 0.96	& 0.97	& 0.96	& 1.00	& 0.98	& 0.97	& 0.98	& 0.97	& 1.00	& 0.97	& 0.97	& 0.96 & 0.97 & 0.97 \\ 
                           & Travel ($VR$) & 0.81	& 0.72	& 0.77	& 0.81	& 0.80	& 0.00	& 0.85	& 0.81	& 0.15	& 0.81	& 0.00	& 0.71	& 0.81	& 0.74 & 0.714 & 0.75 \\ \hline
\multicolumn{2}{c|}{$ACC$}   &\textbf{0.95}	& 0.93	& 0.93	& \textbf{0.95}	& 0.94	& 0.85	& \textbf{0.96}	& 0.95	& 0.86	& \textbf{0.95}	& 0.85	& 0.93	& \textbf{0.95}	& 0.93 & 0.93 & 0.93 \\ \hline
\multicolumn{2}{c|}{$F1$}   &\textbf{0.89}	& 0.84	& 0.86	& \textbf{0.89}	& 0.87	& 0.00	& \textbf{0.92}	& 0.89	& 0.37	& \textbf{0.89}	& 0.00	& 0.84	& \textbf{0.89}	& 0.85 & 0.85 & 0.86\\ \hline
\multicolumn{2}{c|}{Train/Test Time (s)}      & \textbf{16k}	& 26k	& 37k	& \textbf{16k}	& 18k	& 18k	& \textbf{14k}	& 16k	& 20k	& 16k	& \textbf{16k}	 & 16k	& 16k	& \textbf{8k} & 9k & 5k \\ \hline
\end{tabular}
\vspace{-0.1 in}
\end{table*}

\textbf{Encoder-decoder design.} We evaluate six design choices of the encoder-decoder architecture using the \texttt{FU-10K} data set: \# of neural network layers (1), the dropout probability (0), the truncate size on extra-long trajectories (200), the training optimizer (SGD), the use of the attention mechanism (with), and the choice of RNN cells (LSTM). The default setting is given in parentheses. The result in \rtab{ModelParameter} shows that more network layers, the dropout, the other optimizers, and the newer GRU cell do not work better. The attention mechanism does help in our scenario, especially for the recall of travel (more in \rsuba{AttentionVis}). Truncating to smaller segments also improves the performance and also reduces the training time.

\textbf{Model comparison.} On the labeled part of the \texttt{FU-10K} data, \rtab{ModelComparison} lists the performance of all the multiple trajectory inference methods. The encoder-decoder model (ED) is the best in most metrics, with $ACC$ and $F1-ACC$ as high as 0.957 and 0.915. The baseline LSTM gets the second, still more than 10 percent worse than ED in $F1-ACC$. Among other classifiers, DT works the best and achieves a $F1-ACC$ close to 0.7. The other models (LG, Voting, NB, HMM, L-SVM) perform badly in predicting the travels, with a $VP$ or $VR$ smaller than 0.2. Though LG gets the best $SR$, it achieves that by classifying all the records as stay.

\begin{figure*}[t]
\centering
\subfigure[]{\includegraphics[width=1.7 in]{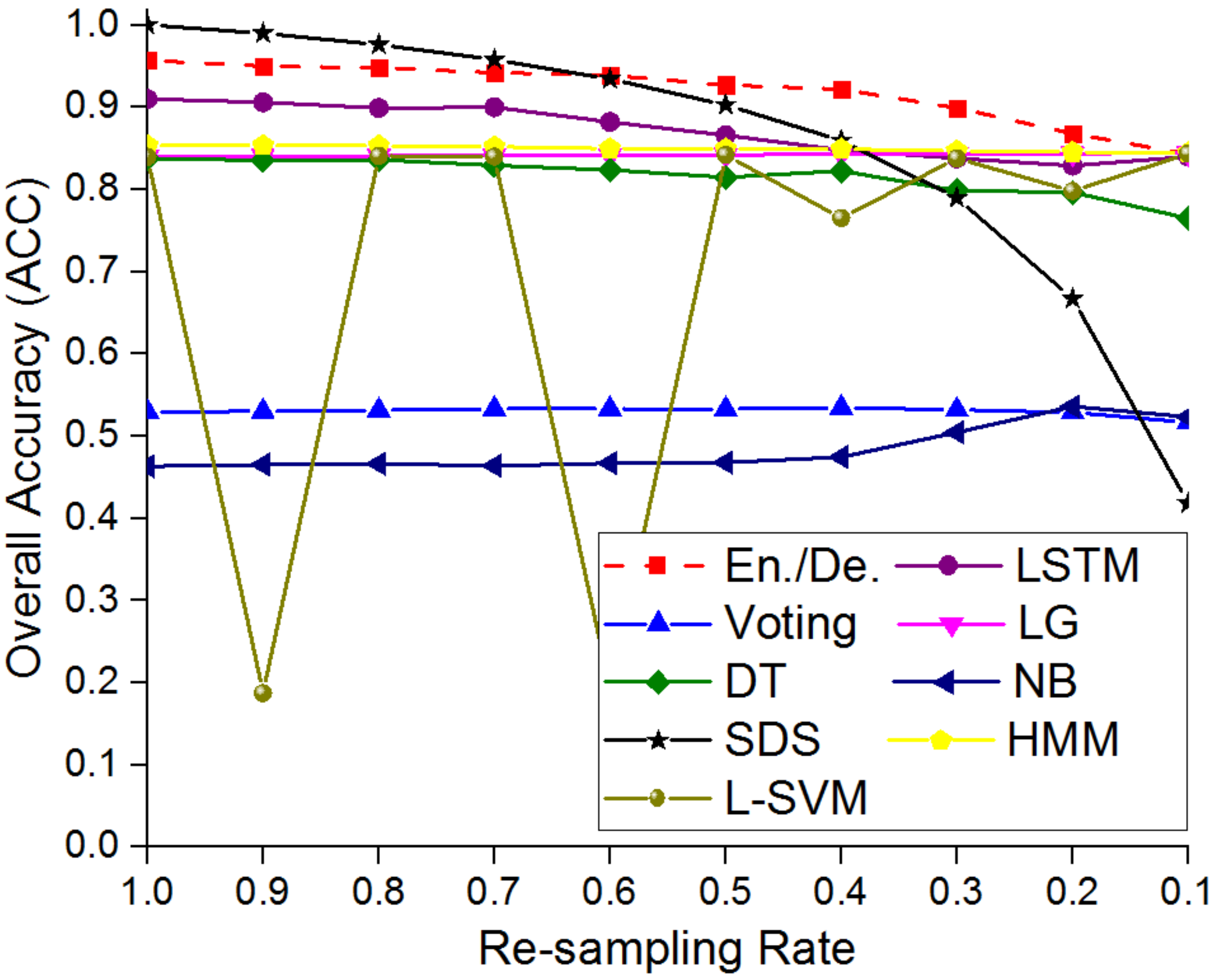}}
\subfigure[]{\includegraphics[width=1.7 in]{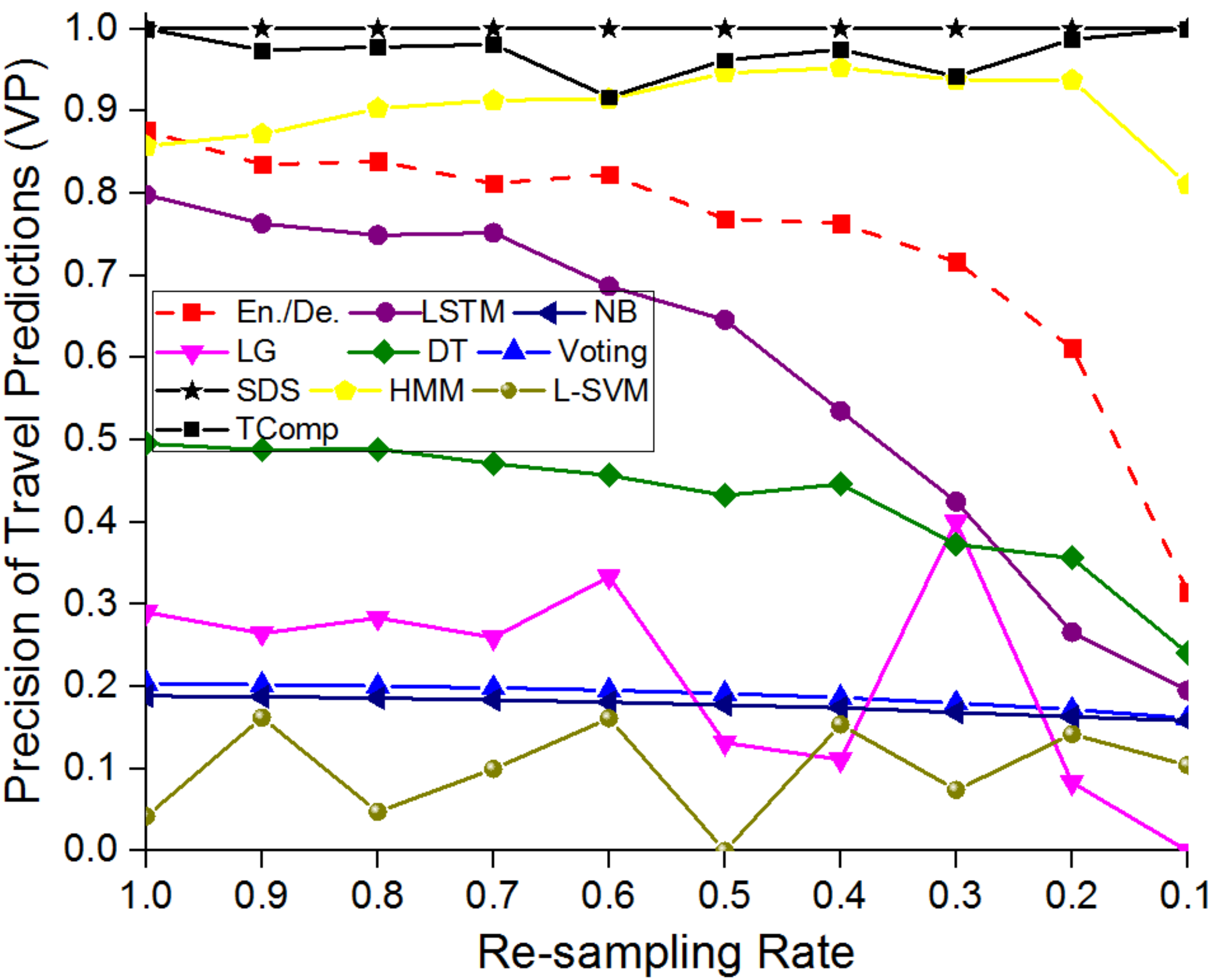}}
\subfigure[]{\includegraphics[width=1.7 in]{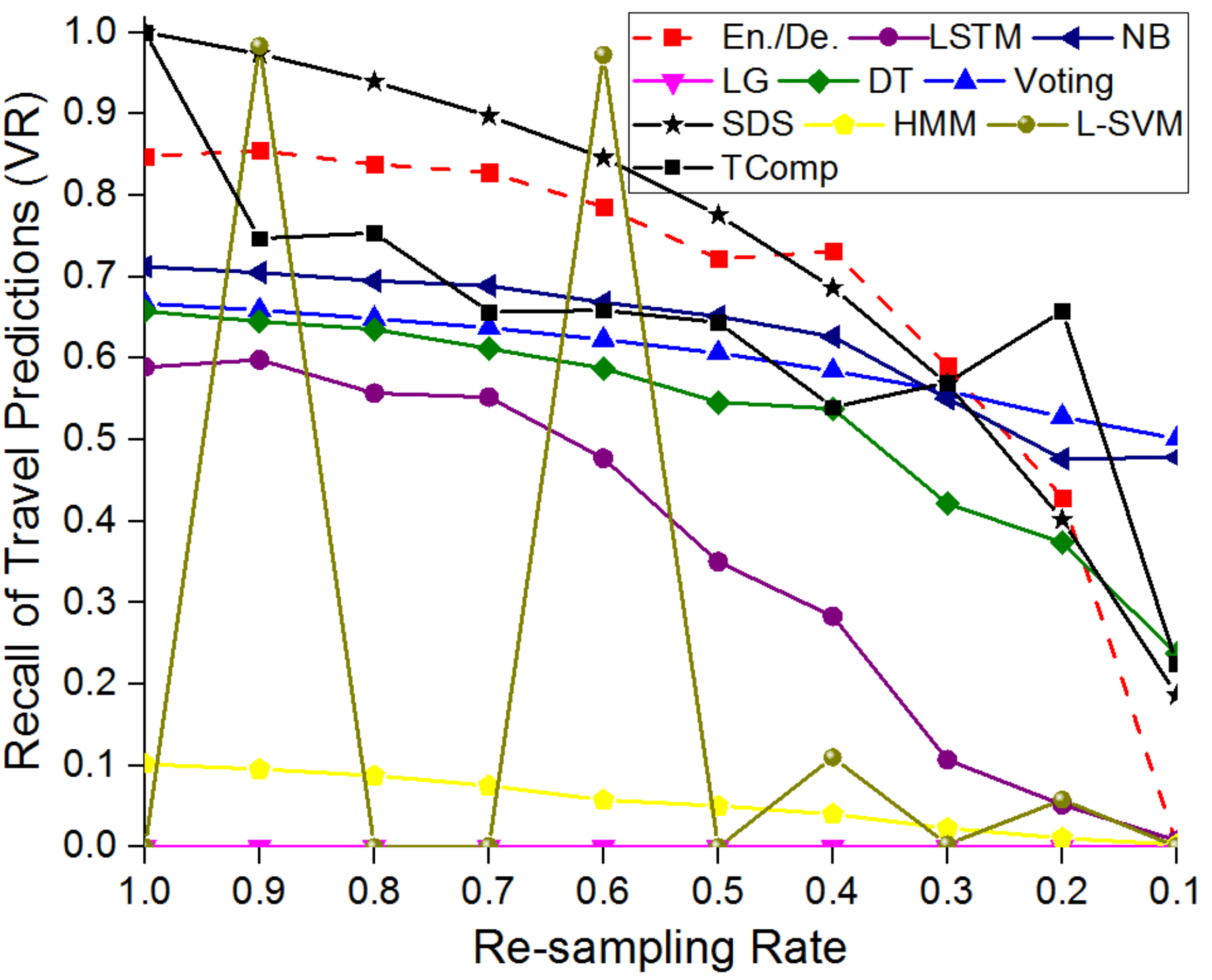}}
\subfigure[]{\includegraphics[width=1.7 in]{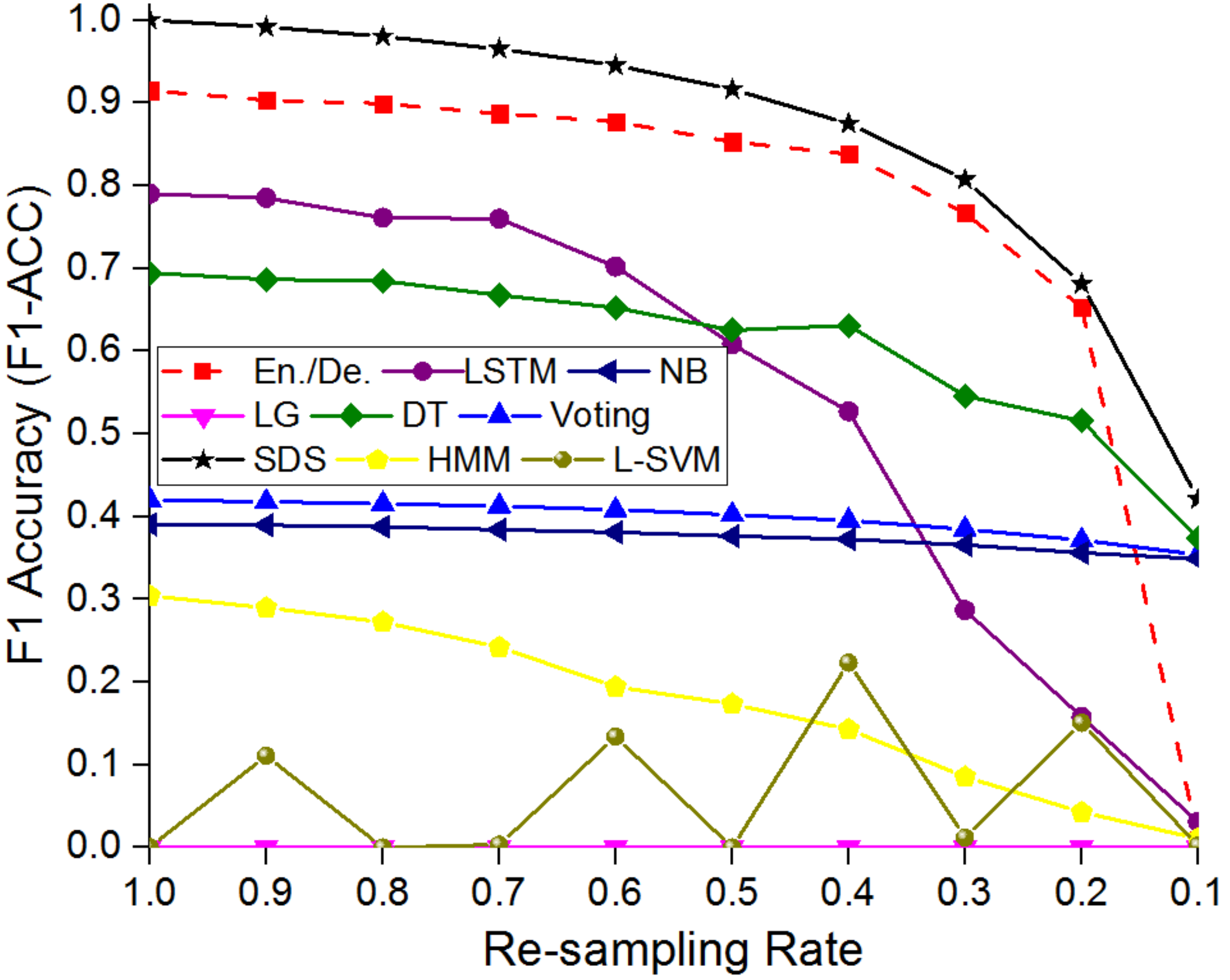}}
\vspace{-0.15 in}
\caption{The comparison of alternative inference methods on \texttt{RE-10K} data. Dashed lines are the proposed encoder-decoder model.}
\vspace{-0.18 in}
\label{fig:ResampledComparison}
\end{figure*}

Extending to the unlabeled part of the \texttt{RE-10K} data, we summarize the performance comparison with different re-sampling rates in \rfig{ResampledComparison}, which also corresponds to an increasing global sparsity (refer to \rfig{SDSPerformance}). Because the stay predictions are generally good for most methods, we only depict $ACC$, $VP$, $VR$, and $F1-ACC$. Note that the performance of the SDS algorithm is also plotted, serving as the upper bound that can be achieved with the single trajectory information. The encoder-decoder is still the best model in most metrics when the re-sampling rate is higher than 0.2, except that HMM has a high precision on a tiny portion of travel records ($VR<0.1$) and L-SVM oscillates between all-stay and all-travel predictions.

Our ED model surpasses the SDS algorithm in $ACC$ from the re-sampling rate of 0.6, starting to enjoy the bonus of the multiple trajectory information. However, it is consistently below the SDS in $F1-ACC$ favoring the travel prediction. According to the theory in \rsec{Labeling}, SDS guarantees a 100\% $VP$, which is much better than the ED model. Starting from a re-sampling rate of 0.3, the travel prediction performance of the ED model dives quickly. Also, trajectory completion \cite{li2016knowledge} does not improve the inference performance. Even worse, because the technique needs to pre-compute a junction network using spatially dense trajectories as input, the test data before completion is constrained into a 5km$\times$5km square region. The precision and recall of SDS in the spatially constrained data set is worse than the randomly sampled \texttt{FU-10K} data.

\begin{figure}[t]
\vspace{-0.1 in}
\centering
\subfigure[]{\includegraphics[width=1.65 in]{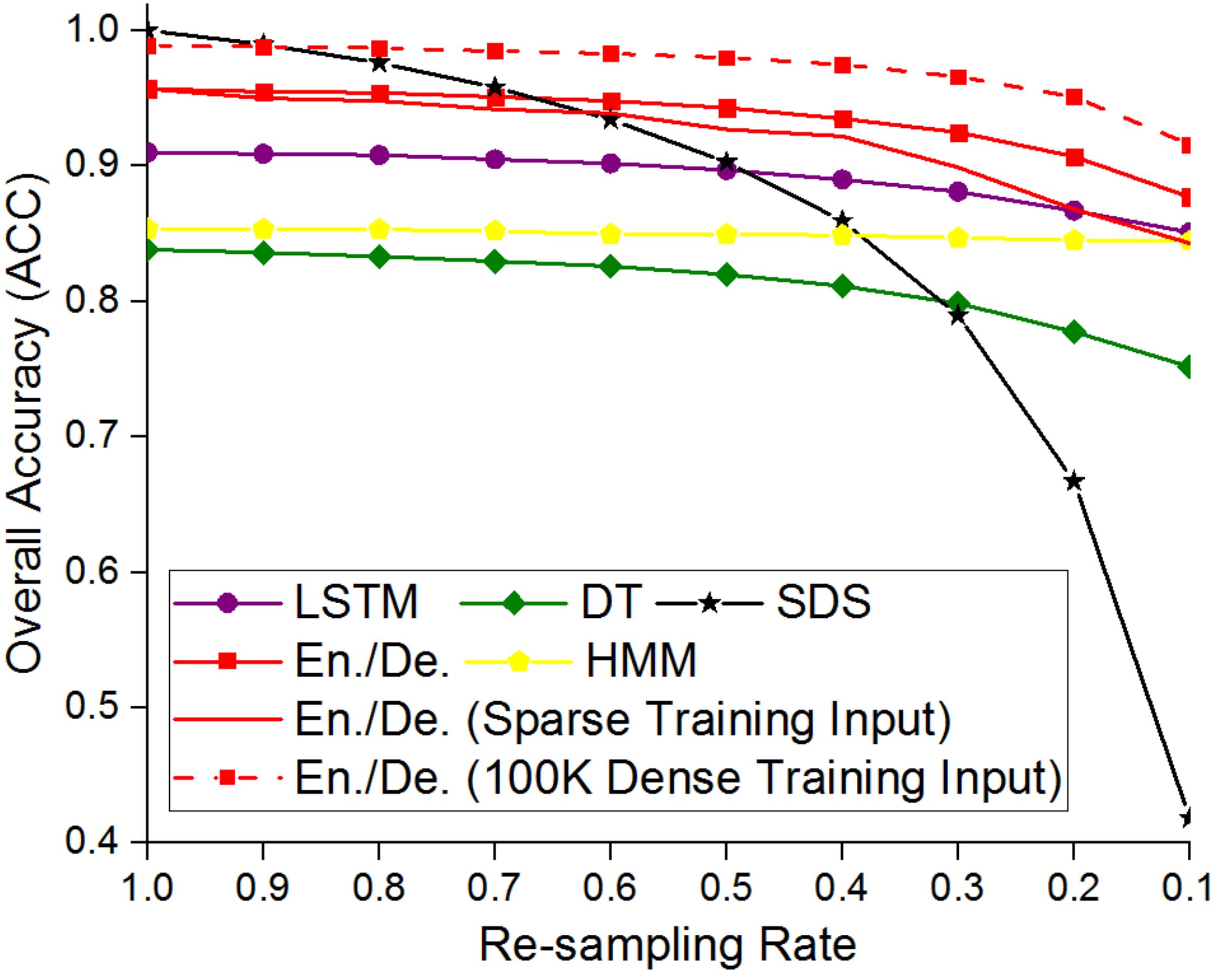}}
\subfigure[]{\includegraphics[width=1.65 in]{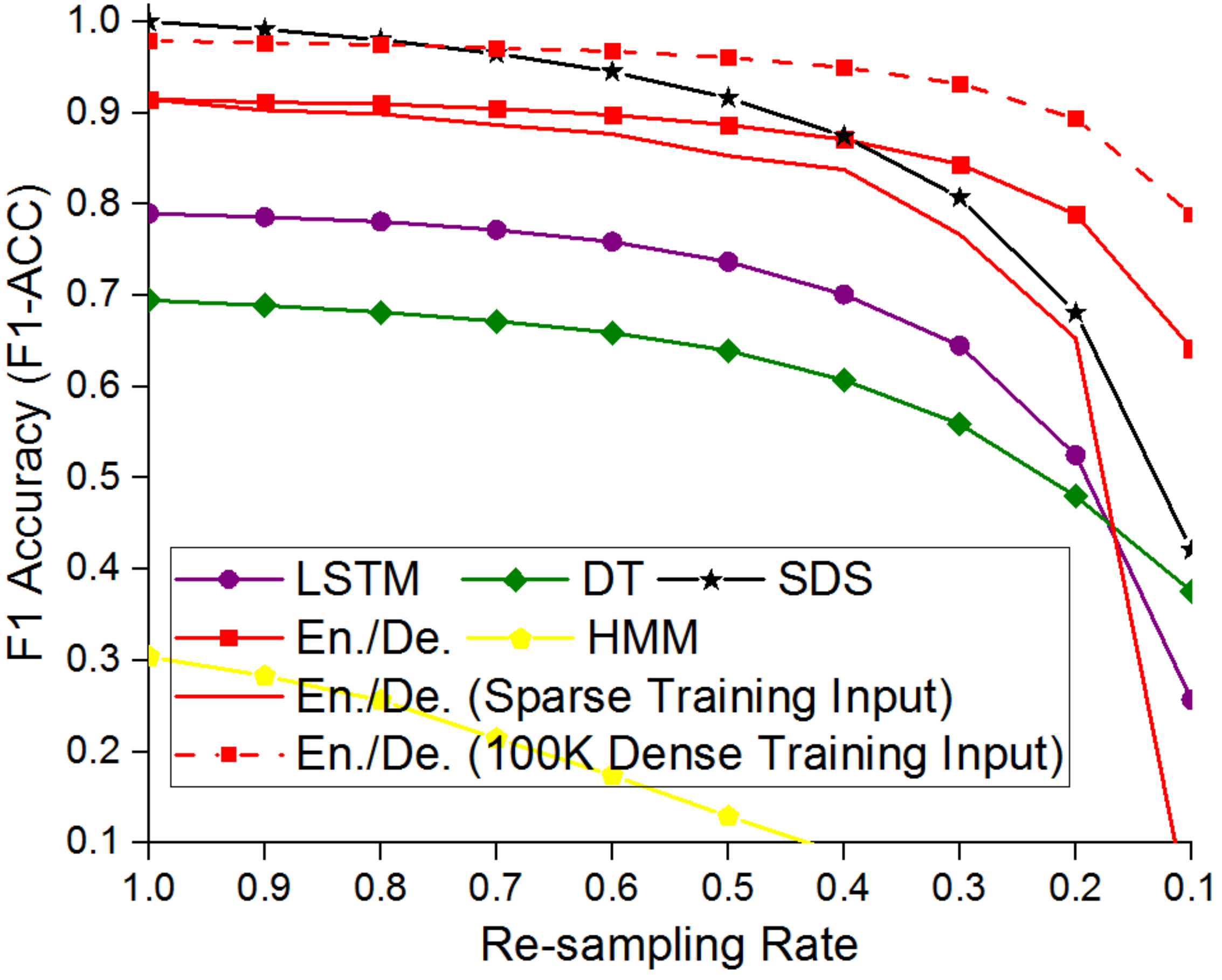}}
\vspace{-0.15 in}
\caption{Training with the 100\% \texttt{RE-10K} data and test on the re-sampled \texttt{RE-10K} data. Red lines are encoder-decoder models.}
\vspace{-0.21 in}
\label{fig:DenseComparison}
\end{figure}

We try to improve the mobility inference by using the densely sampled trajectory as the training data, i.e., the 100\% re-sampled data in \texttt{RE-10K}; and test on the sparse trajectory, i.e., \texttt{RE-10K} with re-sampling rates of 0.1$\sim$1. This is realistic in the model building. As shown in \rfig{DenseComparison}, with denser trajectories and more labels in the training, the test performance of the ED model (straight lines in red with square symbols) is enhanced from the model with sparse input (red lines without symbols), especially in the travel prediction and the re-sampling rate below 0.3. Taking the finding one step further, we use the 100\% re-sampled \texttt{RE-100K} data with 10 times more trajectories for training. The result is surprisingly good -- the ED model outperforms the upper bound of the single trajectory inference from the re-sampling rate of 0.7. In a 10\% re-sampling, ED model achieves $2.19\times~ACC$ and $1.87\times~F-ACC$ compared with SDS.

\begin{table}[t]
\centering
\vspace{-0.01 in}
\small
\caption{Comparison of inference methods on \texttt{FU-10K} data.}
\vspace{-0.1 in}
\label{tab:ModelComparison}
\begin{tabular}{p{.6cm}|p{.55cm}p{.6cm}p{.6cm}p{.55cm}p{.55cm}p{.55cm}p{.55cm}p{.9cm}}
\hline
      & ED              & LSTM   & Voting & LG    & DT    & NB    & HMM   & SVM\\ \hline\hline
$SP$  & \textbf{0.97}  & 0.93 & 0.89 & 0.84  & 0.93  & 0.88 & 0.85 & 0.84 \\ \hline
$VP$  & \textbf{0.88}  & 0.80 & 0.20 & 0.29 & 0.50 & 0.19 & 0.86 & 0.04 \\ \hline
$SR$  & 0.98           & 0.97  & 0.50 & \textbf{1.00} & 0.87 & 0.42 & 1.00 & 1.00 \\ \hline
$VR$  & \textbf{0.85}  & 0.59 & 0.67 & 0.00 & 0.66 & 0.71 & 0.10 & 0.00 \\ \hline
$ACC$ & \textbf{0.96}  & 0.91  & 0.53 & 0.84  & 0.84 & 0.46 & 0.85 & 0.84 \\ \hline
$F1$  & \textbf{0.92}  & 0.79  & 0.42  & 0.00 & 0.69 & 0.39 & 0.30 & 0.00 \\ \hline
\end{tabular}
\vspace{-0.2 in}
\end{table}


\textbf{Scalability and generalizability.} We carry out the same experiment on the full data set with higher numbers of trajectories. As shown in \rtab{Scalability}, the performance keeps steady using the same \texttt{FU-10K} as the training data and test on 10K, 40K, and 100K full trajectory data (\texttt{FU-10K:10K}, etc.). This shows that the model trained on a small data set can be generalized to much larger data sets. Training on the larger data further improves the test performance, which is nearly optimal for \texttt{FU-100K:10K} ($ACC=0.989$, $F1-ACC=0.979$).

\begin{figure}[t]
\centering
\vspace{-0.01 in}
\subfigure[]{\includegraphics[width=1.65 in]{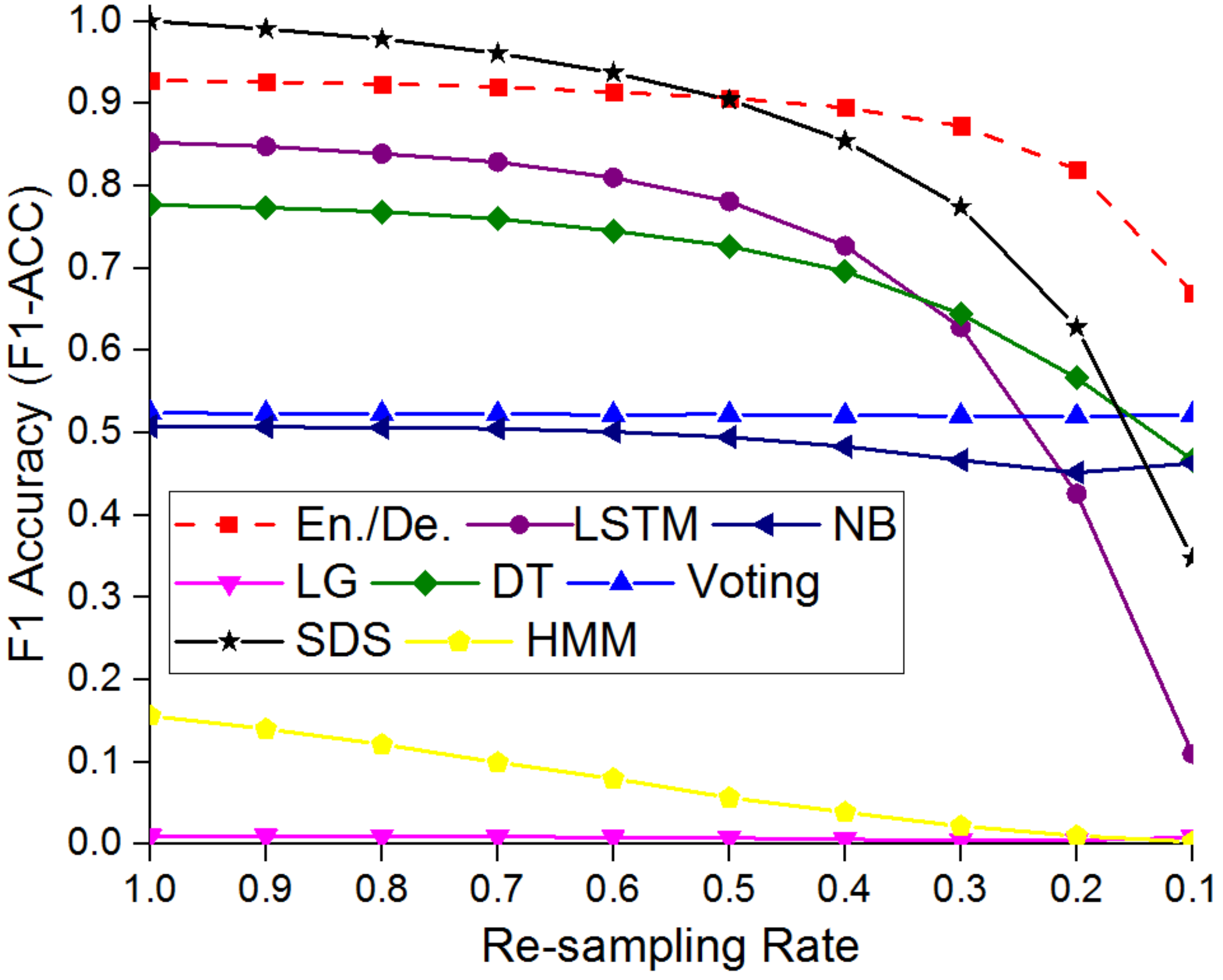}}
\subfigure[]{\includegraphics[width=1.65 in]{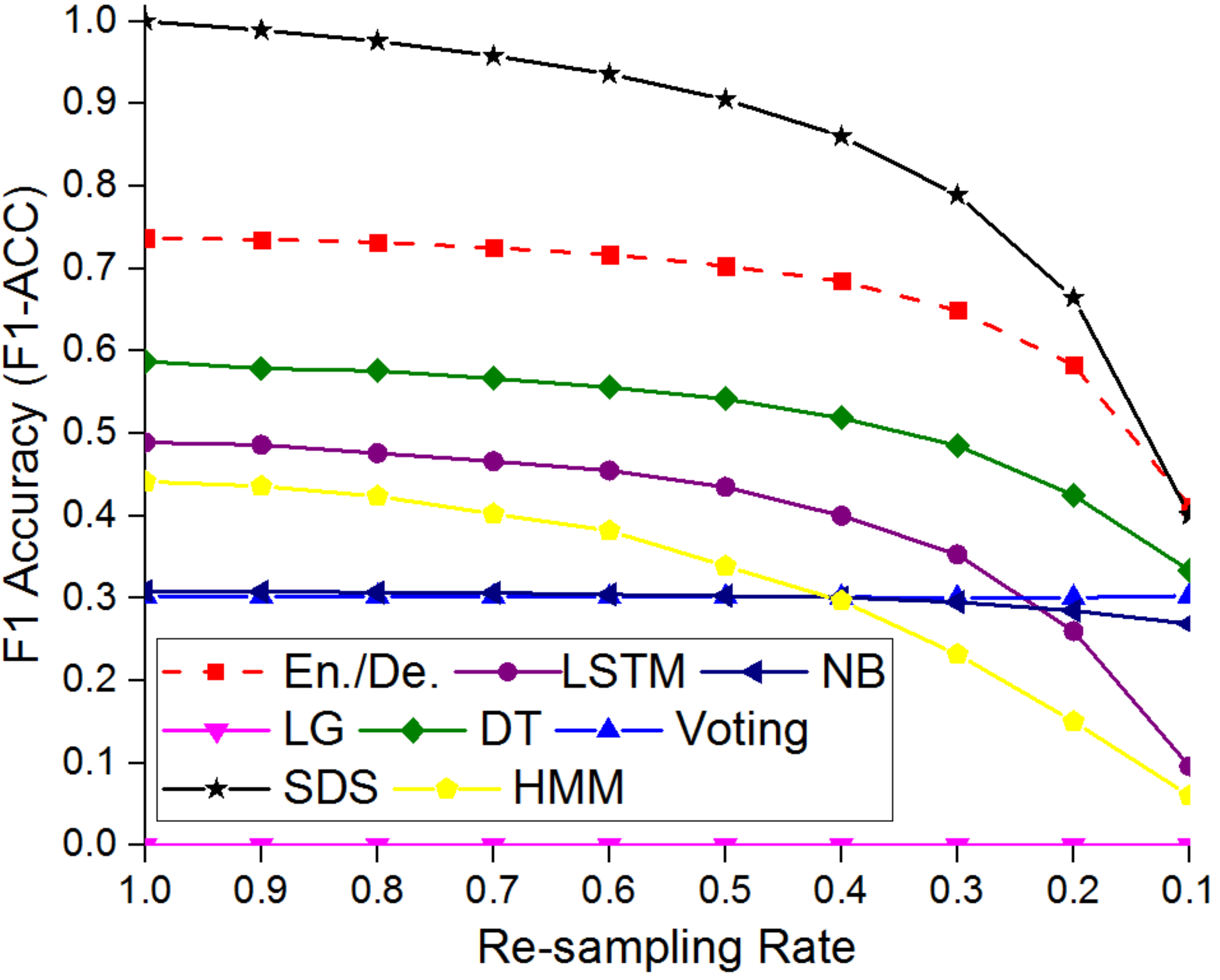}}
\vspace{-0.15 in}
\caption{Experiments with \texttt{RE-10K}: (a) Tianjin; (b) Tangshan.}
\vspace{-0.2 in}
\label{fig:CityDense}
\end{figure}

We conduct the same experiment on \texttt{RE-10K} with the dense trajectory input, using the data sets from Tianjin and Tangshan. Compared with \rfig{DenseComparison}(b) for Beijing, the $F1-ACC$ of the ED model in Tianjin shows a similar curve (\rfig{CityDense}(a)), surpassing the SDS from a re-sampling rate of 0.5. On the other hand, the ED model does not work better than the SDS on the Tangshan data (\rfig{CityDense}(b)). We hypothesize that this is because the selected train/test data in Tangshan has a much lower percentage of travel labels (5.56\%) than Beijing (10.64\%) and Tianjin (16.94\%). The model can not learn the useful pattern given fewer labels. Tangshan is also a smaller city than Beijing and Tianjin, where we have fewer data (\rtab{DataCollection}).

\textbf{Implications.} The experiment result demonstrates that the SDS algorithm is accurate on the single trajectory (100\% $SP$ and $VP$). The optimized encoder-decoder models can learn from the multiple trajectory input to improve the single trajectory mobility inference through the excellent generalizability to sparse trajectories and the scalability to large training data. In fact, we expect the proposed model to perform even better in comparison to the SDS algorithm. We only evaluate on the part of the trajectory labeled in the 100\% re-sampled test data. For the unlabeled test data (38.9\% for Beijing), it is reasonable to guess that our model performs similarly to the labeled part, while the SDS can not infer at all. In future, we plan to develop the re-sampling mechanism on the simulated data to test on the 100\% labels of the trajectory data set.

\bsec{Related Work}{Related}

\bsubsec{The Study of Urban Trajectory}{Rel-Trajectory}

Using the trajectory data in the city to understand the urban activity and human mobility has been a recent focus of study \cite{zheng2015trajectory}.
On the continuously measured trajectory, the detailed route information is available for analysis \cite{LiuAR10}\cite{liu2011diverse}\cite{yuan2013t}. For instance, the trajectories of taxis can be used to classify drivers by their job performance \cite{LiuAR10}, or aggregated as time-dependent landmark graph \cite{yuan2013t} and trajectory visualization \cite{liu2011diverse}, in order to compute the fastest route for drivers.
Based on over one million bank note circulation reports in US, Brockmann et al. explained the human mobility as the combination of a scale-free jump and a heavy-tailed wait, and proposed a random-walk model to characterize these findings \cite{brockmann2006scaling}. 
The group of Barab\'asi explained the high degree of spatiotemporal regularity in human mobility by the tendency to avoid visiting new places and to return to the previously visited locations \cite{gonzalez2008understanding}\cite{song2010modelling}. 

On the analysis of urban trajectories, the need for separating stay and travel has been partially met by the greedy algorithms similar to \ralg{Optimal} \cite{calabrese2010geography}\cite{jiang2013review}. Nevertheless, none of these works formally define the stay/travel state of a trajectory, nor do they consider the mobility inference problem on sparse trajectories.

\begin{table}[t]
\small
\setlength{\tabcolsep}{2pt}
\centering
\caption{The scalability of the Encoder-Decoder architecture.}
\vspace{-0.08 in}
\label{tab:Scalability}
\begin{tabular}{c|cccccc}
\hline
                  & \texttt{10K}  & \texttt{10K:40K}  & \texttt{10K:100K} & \texttt{100K:10K} & \texttt{100K:40K}  & \texttt{100K:100K}   \\ \hline\hline
$SP$                & 0.97	& 0.97	& 0.97	& \textbf{0.99}	& 0.99	& 0.99 \\ \hline
$VP$                & 0.88	& 0.88	& 0.88	& \textbf{0.97}	& 0.97	& 0.97 \\ \hline
$SR$                & 0.98	& 0.98	& 0.98	& \textbf{0.99}	& 0.99	& 0.99 \\ \hline
$VR$                & 0.85	& 0.85	& 0.85	& 0.96 & 0.96	& \textbf{0.96} \\ \hline
$ACC$               & 0.96	& 0.96	& 0.96	& \textbf{0.99}	& 0.99	& 0.99 \\ \hline
$F1$            & 0.92	& 0.91	& 0.91	& \textbf{0.98}	& 0.98	& 0.98 \\ \hline
Time (s)          & \textbf{17k} & 18k & 18k & 150k & 150k & 157k \\ \hline
\end{tabular}
\vspace{-0.16 in}
\end{table}

\vspace{-0.06 in}

\bsubsec{The Inference of Sparse Trajectory}{Rel-Sparse}


There are two definitions of the sparse trajectory in the literature of urban data analysis. The first one considers a sequence of infrequent reports from vehicles. 
These trajectories are usually collected in a uniform time interval of seconds or a few minutes. 
We call them the temporally sparse trajectory \cite{herring2010estimating}\cite{rahmani2013path}\cite{rahmani2012path}\cite{sanaullah2016developing}\cite{li2017citywide}. The second class is the spatially sparse trajectory data in which many road segments in a city are not covered by any of the trajectory, especially for a given period of time. 
The literature on this class of data mostly studied the travel time estimation problem \cite{wang2014travel}\cite{sanaullah2016developing}\cite{li2017citywide}.


We mainly consider the mobility inference problem on the temporally sparse trajectory, as our data set covers most of the city regions. 
The recent works on this topic focus on the extraction of travel paths \cite{rahmani2013path}\cite{rahmani2012path}\cite{li2017citywide}. Typically, the problem is decomposed into two tasks: the map-matching and the path-inference. In map-matching, each location record on the trajectory is matched to a point on a particular segment of the road network \cite{quddus2007current}\cite{ochieng2003map}. In path-inference, the matched points on the map are connected by shortest paths to form the travel path \cite{lou2009map}\cite{he2013line}. 


The map-matching based techniques can not be applied directly to our mobility inference problem. First, the trajectory data in our case encompasses not only the movement of high-speed vehicles on the ground, but also those by bikes and subways. The locations of these trajectories may not be on the road network, thus are not appropriate for map-matching. It is also costly to evolve the technique with the fast-changing road network of modern cities. Second, we have both travel and stay in our data while the previous approaches mostly work on the travel part of the trajectory with a temporal sparsity two orders of magnitudes smaller than our case.
The trajectory completion techniques can also be applied to compute dense trajectories from known sparse ones. In Ref. \cite{li2016knowledge}, Yang et al. proposed a geometry-based method that pre-computes the junction networks in cities and then predicts the missing part of the travel trajectory, without knowing the city map. However, their technique requires the speed and heading information of each location record, and spatially dense trajectory data set to pre-compute the junction network. Applying the same trajectory completion technique on our problem leads to worse performance than the proposed deep learning method.



\bsubsec{Deep Learning for Urban Analytics}{Rel-DL}


Many deep learning methods have been customized to work with the urban data. Lv et al. proposed a stacked autoencoder architecture to learn the generic traffic flow features from the data collected at road-side detectors \cite{lv2015traffic}. Zhang et al. designed standalone residual networks to model three key temporal properties of crowd flows and dynamically aggregated the network outputs to predict the inflow and outflow of traffic in city regions \cite{zhang2017deep}. Yao et al. presented DeepSense \cite{yao2017deepsense}, a deep learning framework to resolve the data noise and extract useful features from the mobile sensing data.


Yet, the sequence to sequence neural network models (e.g., LSTM \cite{LSTM}) are rarely used in the urban data analytics. The recent work by Zhao et al. on the next location recommendation from time-aware trajectory data \cite{zhao2017time} comes the closest to our study, though there are fundamental differences. We proposed an end-to-end neural network model to better understand the current trajectory and make inference at any spatiotemporal records. In contrast, Zhao et al. targeted at the one-time prediction of the next activity in the future. Their prediction is based on a ranking function with the deep learning method only used for embedding.

\bsec{Conclusion}{Conc}

This paper studies the problem of mobility inference over sparse trajectories. Based on the observation of a long-tailed sparsity pattern in the trajectory data, we design a single trajectory inference algorithm that detects the mobility of close to half of trajectory records with a guaranteed 100\% precision. Furthermore, we propose an encoder-decoder architecture that learns the mobility pattern from multiple trajectories. The learning model significantly outperforms the traditional classifiers in all the performance metrics. In particular, by feeding with the large-scale densely sampled training data, our model achieves a near-optimal overall accuracy on the records labeled by the single trajectory inference algorithm. On unlabeled records, our model outperforms the single trajectory inference by a factor of two on extremely sparse trajectories. Experiment results also demonstrate that our model generalizes to different urban data sources and scales to large data sets.

\bibliographystyle{IEEEtran}
\bibliography{ICDE20}

\appendices

\bsec{Proofs and the Exact Algorithm}{Proof}

\textsc{Theorem 1: Intrinsic linkage between discrete and continuous mobility of a trajectory.}

\bproof
\rthe{Continuous-Discrete}(a). For the discrete stay segment $\gamma$ in the time series $\omega = \{t_p,\cdots,t_q\}\;(t_q-t_p \geq \Delta{T})$, consider its corresponding continuous segment $\gamma'$ in the time period $\tau = [t_p,t_q]$. $\gamma'$ satisfies $|\tau| = t_q-t_p \geq \Delta{T}$. $\forall t_i,t_j \in \tau$, we have $||\ell(t_i)-\ell(t_j)|| \leq ||\ell(t_i)-\ell(t_i')|| + ||\ell(t_i')-\ell(t_j')|| + ||\ell(t_j')-\ell(t_{j})||$, given that the straightline is the shortest distance between $\ell(t_i)$ and $\ell(t_j)$. Here $t_i'$ and $t_j'$ are the closest time point in $\omega$ to $t_{i}$ and $t_{j}$ respectively. Because $||\ell(t_i')-\ell(t_j')||<\Delta{S}$, $||\ell(t_i)-\ell(t_i')|| \leq \epsilon \cdot v_{max}$, $||\ell(t_j')-\ell(t_{j})|| \leq \epsilon \cdot v_{max}$, we have $||\ell(t_i)-\ell(t_j)||< \Delta{S} + 2 \cdot \epsilon \cdot v_{max}$. That is, $\gamma'$ is a stay segment by \rdef{Continuous-Segment}(a) under the parameters of $\Delta'{S} = \Delta{S} + 2 \cdot \epsilon \cdot v_{max}$ and $\Delta'{T} = \Delta{T}$.

\rthe{Continuous-Discrete}(b). For the discrete travel trip $\gamma$ in the time series $\omega = \{t_p,\cdots,t_q\}$, by definition, we have $\forall t_i', t_j' \in \omega\;(t_j' - t_i' \geq \Delta{T})$, there exist two time points $t_m', t_n' \in \omega\;(t_i' \leq t_m' < t_n' \leq t_j')$ satisfying $||\ell(t_m')-\ell(t_n')|| \geq \Delta{S}$. Consider the corresponding continuous segment $\gamma'$ in the time period $\tau = [t_p,t_q]$, $\forall t_i, t_j \in \tau\;(t_j - t_i \geq \Delta{T} + 2 \cdot \epsilon)$, we can find $t_i'$ (the closest time point in $\omega$ no smaller than $t_i$) and $t_j'$ (the closest time point in $\omega$ no larger than $t_j$), having $t_j' - t_i' \geq \Delta{T}$. There exist two time points $t_m', t_n' \in \omega\;(t_i \leq t_i' \leq t_m' < t_n' \leq t_j' \leq t_j)$ satisfying $||\ell(t_m')-\ell(t_n')|| \geq \Delta{S}$. That is, $\gamma'$ is a travel trip by \rdef{Continuous-Segment}(b) under the parameters of $\Delta'{S} = \Delta{S}$ and $\Delta'{T} = \Delta{T} + 2 \cdot \epsilon$.
\eproof

\vspace{-0.45cm}

\noindent \textsc{Theorem 2: Continuous mobility of dense stay segments.}

\bproof
\rthe{Continuous-Dense}(a). For the dense stay segment $\gamma$ defined in the time series $\omega = \{t_p,\cdots,t_q\}$, consider its corresponding continuous segment $\gamma'$ in the time period $\tau = [t_p,t_q]$. We have $|\tau| = t_q-t_p \geq \Delta{T}$ because $\gamma$ is the dense stay segment. For any two time points $t, t' \in [t_p, t_q]$, denote the closest time points in the time series of $\omega$ to $t$ and $t'$ as $t_i$ and $t_j$ ($p \leq i \leq q,\;p \leq j \leq q$). We have $||\ell(t)-\ell(t')|| \leq ||\ell(t)-\ell(t_i)|| + ||\ell(t_i)-\ell(t_j)|| + ||\ell(t_j)-\ell(t')|| < \Delta{S}/3 + \Delta{S}/3 + \Delta{S}/3 = \Delta{S}$ by \rprop{ContinuousAssumption}. The conditions for the continuous model of the stay segment in \rdef{Continuous-Segment}(a) then hold.

\rthe{Continuous-Dense}(b). For the discrete segment $\gamma$ defined in the time series $\omega = \{t_p,\cdots,t_q\}$, consider its corresponding continuous segment $\gamma'$ in the time period $\tau = [t_p,t_q]$. If $\exists p \leq i < q,\;t_{i+1}-t_{i} > \Delta{T}$, i.e., the unobserved time period of $(t_{i}, t_{i+1})$ has a duration longer than $\Delta{T}$. Observing a time period $\tau' \subset (t_{i}, t_{i+1})$ with $|\tau'| = \Delta{T}$ can detect a different stay segment from the other part of the segment in $(t_{i}, t_{i+1})$. Then there can be travel trips surrounding the segment in $\tau'$ to connect the trajectory. This possibility can not be validated or rejected given the information of the discrete segment $\gamma$ only. Therefore, the corresponding continuous segment $\gamma'$ can not be inferred as stays, unless $\forall p \leq i < q,\;t_{i+1}-t_{i} \leq \Delta{T}$.

On the dense segment $\gamma$, if the corresponding continuous segment $\gamma'$ is the stay segment, by \rdef{Continuous-Segment}(a), $\forall t_i,t_j \in \omega \subset \tau$, $||\ell(t_i)-\ell(t_j)|| < \Delta{S}$. Therefore, $\gamma$ must be a dense stay segment.
\eproof

\vspace{-0.45cm}

\begin{algorithm}[t]
 \SetEndCharOfAlgoLine{}
\SetKwInOut{Input}{Input}\SetKwInOut{Output}{Output}
\newcommand\mycommfont[1]{\footnotesize{#1}}
\SetCommentSty{mycommfont}
\newcommand\mylnfont[3]{\footnotesize{#1}}
 \Input{$\Gamma=\bigcup_{i \in [1,L]}<t_i,\ell(t_i)>, t_1<\cdots<t_L$ (dense trajectory), $\Delta{T}$, $\Delta{S}$ (the space and time parameters)}
 \Output{$I_{S/T}(t_i), \forall i \in [1,L]$ (the mobility of each record)}

 \Begin{
 \For{$head \leftarrow [1,L-1]$}{
     \For{$cursor \leftarrow [head+1,L]$}{
         \tcc*[l]{iterate all the candidate stay segments}
         \If{$t_{cursor}-t_{head} \geq \Delta{T}$}{
             \For{$i \leftarrow [head,cursor-1]$}{
                 \For{$j \leftarrow [i+1,cursor]$}{
                     \If{$||\ell(t_{i})-\ell(t_{j})|| \geq \Delta{S}$}{
                         $Stay \leftarrow$ False,~\Break
                     }
                 }
             }
             \If{$Stay!=$False}{
                 \For{$i \leftarrow [head, cursor]$}{
                    $I_{S/T}(t_i) \leftarrow$ S
                 }
             }
         }
     }
 }
 \tcc*[l]{the remaining records are travel trips}
 \For{$i \leftarrow [1, L]$}{
     \If{$I_{S/T}(t_i) !=$ S}{
         $I_{S/T}(t_i) \leftarrow$ T
     }
 }
 \Return $I_{S/T}(t_i), i=[1,L]$
 }
\caption{The exact algorithm on dense trajectories.}
\label{alg:Optimal}
\vspace{-0.05 in}
\end{algorithm}

\noindent \textsc{Theorem 3: Continuous mobility of travel records.}

\bproof
\rthe{Continuous-Travel}(a). For the record at time $t_i$, consider any time period $\tau = [t, t']$ satisfying $|\tau| \geq \Delta{T}$ and $t_i \in \tau$. If the three conditions hold, the time period of $\tau'=[t_p, t_q]$ satisfies $|\tau'| \leq \Delta{T}$ and $t_i \in \tau'$. We have $t_p \in \tau$ or $t_q \in \tau$. Otherwise, we will have $t_p < t$ and $t_q > t'$, which leads to the contradiction of $|\tau'| = t_q - t_p > t' - t = |\tau| \geq \Delta{T}$. For the either case of $t_p \in \tau$ or $t_q \in \tau$, we have $||\ell(t_i)-\ell(t_{p})|| \geq \Delta{S}$ and $||\ell(t_i)-\ell(t_{q})|| \geq \Delta{S}$. This contradicts to \rdef{Continuous-Segment}(a). Therefore, the record at time $t_i$ can not be in any stay segment, and it must be in a travel trip by \rdef{Continuous-Segment}(b).

\rthe{Continuous-Travel}(b). For the record at time $t_i\;(1<i<L)$, if the condition does not hold, $\forall 1 \leq p<i<q\leq L$ satisfying $t_q-t_p \leq \Delta{T}$, we have $||\ell(t_i)-\ell(t_{p})|| < \Delta{S}/2$ or $||\ell(t_i)-\ell(t_{q})|| < \Delta{S}/2$.

Consider the smallest time point $t_j$ satisfying $t_j>t_i$ and $||\ell(t_i)-\ell(t_{j})|| \geq \Delta{S}/2$. We should have $t_{j} - t_{i} \leq \Delta{T}$ because otherwise $\forall t_i<t_k<t_j,\;||\ell(t_i)-\ell(t_{k})|| < \Delta{S}/2$. There exists a time period of $\tau'=[t_i,t_j)$, for all observed $t_k \in \tau'$, $||\ell(t_i)-\ell(t_{k})|| < \Delta{S}/2$. Then $\forall t_{k}, t_{k'} \in \tau'$, $||\ell(t_k)-\ell(t_{k'})|| \leq ||\ell(t_i)-\ell(t_{k})|| + ||\ell(t_i)-\ell(t_{k'})|| < \Delta{S}$, $t_i$ will be possibly in a stay segment, without the information to reject the possibility.

Having $t_{j} - t_{i} \leq \Delta{T}$ and $||\ell(t_i)-\ell(t_{j})|| \geq \Delta{S}/2$, using the proof by contradiction, we have $\forall k<i$ satisfying $t_j-t_k \leq \Delta{T}$, we have $||\ell(t_i)-\ell(t_{k})|| < \Delta{S}/2$. Consider the largest time point $t_{j'}$ satisfying $t_j-t_{j'} > \Delta{T}$, we can construct a time period of $(t_{j'},t_j)$, for all the observed time point of $t_k$ having $j' < k < j$, we have $||\ell(t_i)-\ell(t_{k})|| < \Delta{S}/2$. The distance between these observed time points is below $\Delta{S}$. Then there can be a continuous segment in the time period of $\tau' \subset (t_{j'},t_j)$ satisfying $|\tau'|=\Delta{T}$. We do not have any information to reject the inference of stays on this segment. Therefore, the record at $t_i \in \tau'$ can not be in any travel trip.
\eproof

\vspace{-0.45cm}

We introduce an exact algorithm to infer the discrete mobility (\rdef{Discrete-Segment}) from densely sampled trajectories, as shown in \ralg{Optimal}. The algorithm iterates all the candidate segments in a trajectory to decide whether they meet the condition of stays. The records not in any stay segments are travels. The algorithm has a computational complexity of $O(L^4)$ ($L$ is the number of records in a trajectory), which is computationally infeasible for the large-scale trajectory data. In our targeted scenario, we do not have the densely sampled trajectory.

\bsec{Material for the SDS algorithm}{Proposition}

\begin{figure}[t]
\centering
\includegraphics[height=1.8 in]{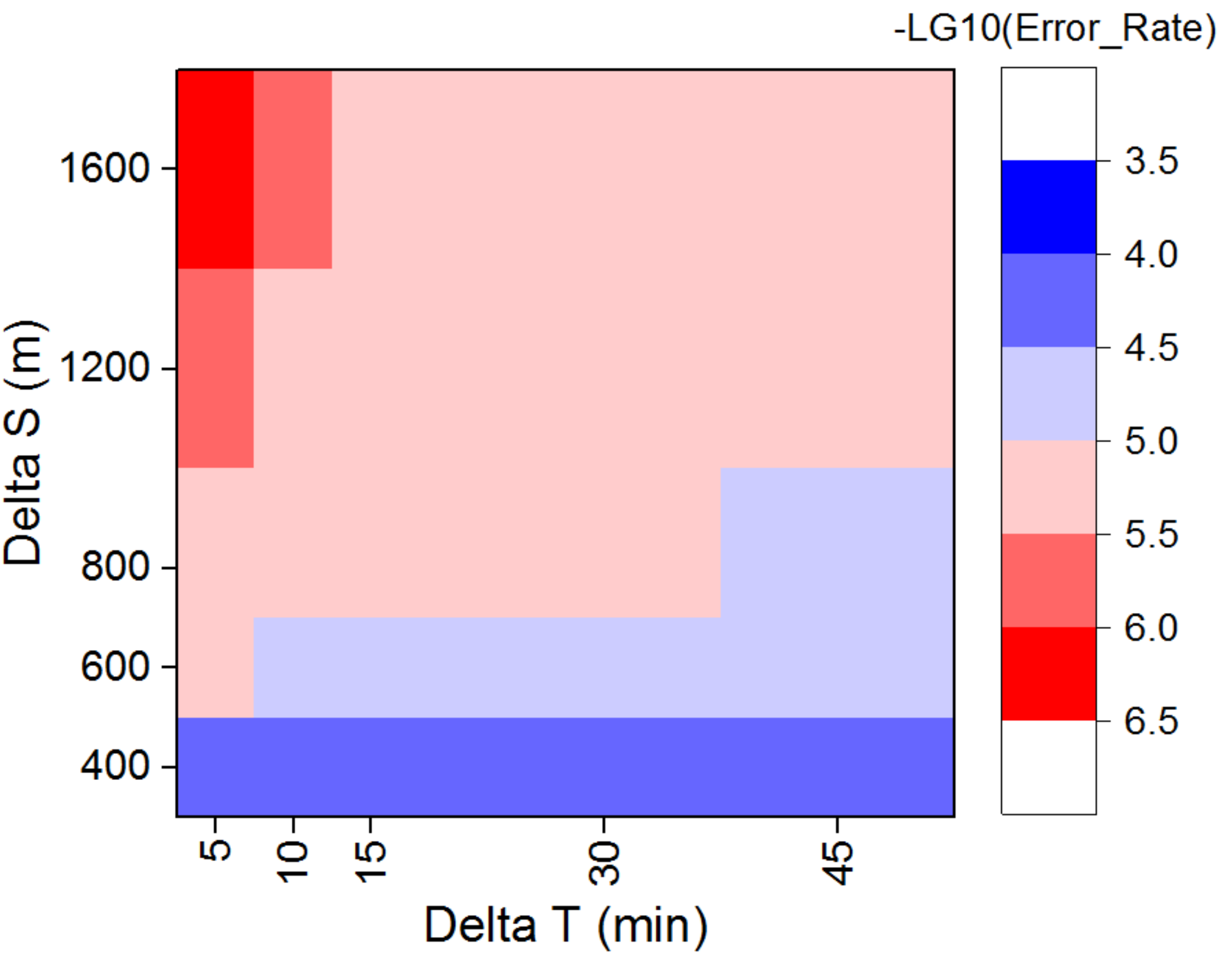}
\vspace{-0.1 in}
\caption{The probability for violating Observation 1, under different $\Delta{S}$ and $\Delta{T}$, mapped by the $-\lg_{10}$ operator.}
\vspace{-0.2 in}
\label{fig:Proposition}
\end{figure}

\begin{figure}[t]
\centering
\subfigure[]{\includegraphics[width=1.6 in]{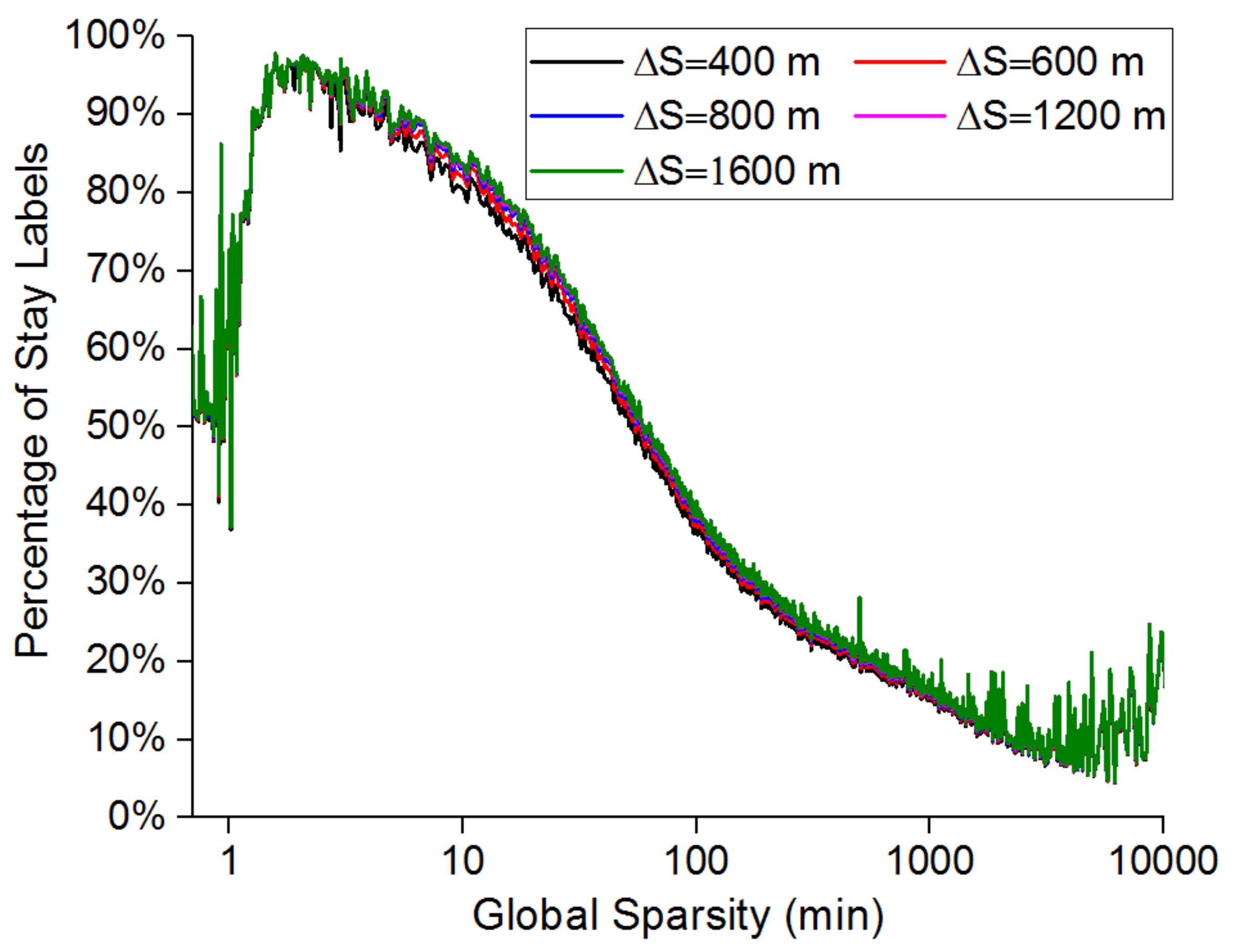}}
\subfigure[]{\includegraphics[width=1.6 in]{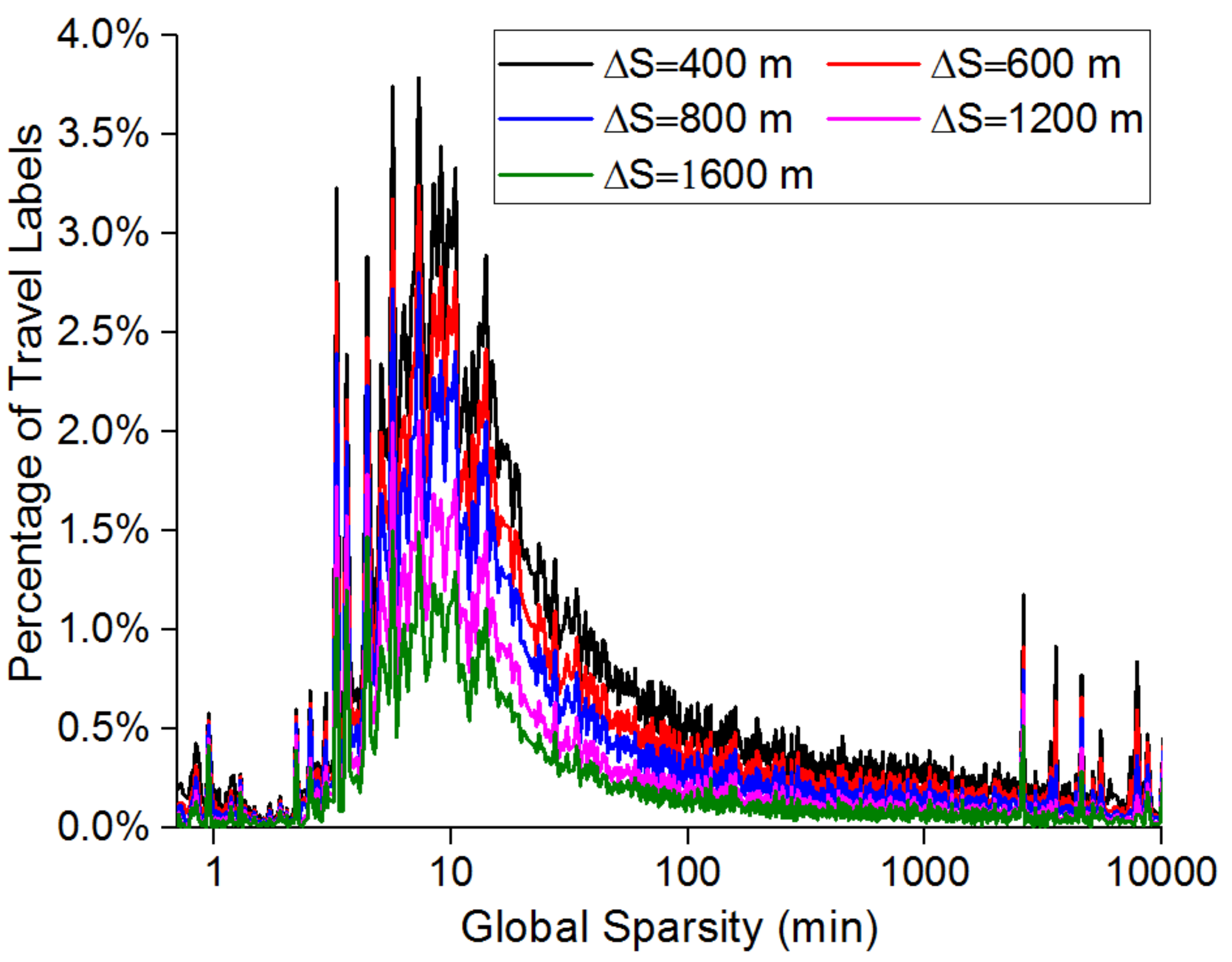}}\\
\vspace{-0.1 in}
\caption{The percentage of stay/travel on the trajectory with different global sparsity values and $\Delta{S}$ ($\Delta{T}$ fixed to 30 min).}
\vspace{-0.15 in}
\label{fig:TrajectoryLabeling2}
\end{figure}

To validate \rprop{ContinuousAssumption}, we conduct an experiment on the full data set in Beijing. For each record in a trajectory, we explicitly remove the record and detect all the dense stay segments from the remaining trajectory. If the record is within a dense stay segment, we check whether the record, as time $t$, violates \rprop{ContinuousAssumption}. As shown in \rfig{Proposition}, among 10-billion potential $<$record, interval$>$ pairs for each parameter setting, the probability of violating \rprop{ContinuousAssumption} is below $10^{-5}$ if $\Delta{S} \geq 800 m$ and $\Delta{T} \leq 30 min$.

In the mobility definition of the trajectory model, the parameters of $\Delta{S}$ and $\Delta{T}$ need to be determined. In fact, these parameters provide the flexibility to capture the multi-scale mobility in the human trajectory. Inside the city boundary, $\Delta{T}$ and $\Delta{S}$ can be minutes and meters to describe the short-term stays and travels; while in the state level, $\Delta{T}$ and $\Delta{S}$ can be days and hundreds of kilometers to characterize the stay in a city and the travel between cities.

We focus on the detection of intra-city travels because the number of travel records is much fewer than the stay and the detection of stay is relatively insensitive to the parameter change (\rfig{TrajectoryLabeling}(a), \rfig{TrajectoryLabeling2}(a)). The goal is to detect more travels while keeping the mobility definition reasonable. According to \rfig{TrajectoryLabeling}(b), we pick $\Delta{T}=30 min$ because the detected ratio of travel does not increase much when switching to $\Delta{T}=45 min$ and it does not impose a strict stay definition which violates \rprop{ContinuousAssumption}. Similarly, according to \rfig{TrajectoryLabeling2}(b), we pick $\Delta{S}=800 m$ which maximizes the recall of travel ($\frac{SDS(\Gamma, T, \Delta{S}, \Delta{T})}{SDS(\Gamma, T, \Delta{S}/2, \Delta{T})}$) and allows a mild stay definition compared with $\Delta{S}=400 m$. The parameters of $\Delta{T} = 30 min$ and $\Delta{S} = 800 m$ are consistent with the empirical settings in Ref.  \cite{calabrese2010geography}\cite{jiang2013review}.

\bsec{Generation of the simulation data}{Simulation}

First, we apply the CTRW model in \cite{brockmann2006scaling}\cite{gonzalez2008understanding} to generate artificial human trajectories. The model characterizes the human trajectory as a two-state interplay between the scale-free displacements (travel) and a long-tailed waiting time distribution (stay). The probability density functions of both the travel distance and the waiting/stay time apply the truncated power-law function observed in \cite{gonzalez2008understanding}. The exponent parameters of the function are calibrated by our trajectory data set applying the SDS algorithm. Each trajectory starts from a random location. Within each stay period, the location at any time is computed by the stay location plus a random spatial offset smaller than $\Delta{S}/2$. The travel between consecutive stay locations is assumed to be a straight-line, constant-speed trajectory. The parameter speed controls the ratio of the stay/travel time.

Second, each trajectory is sampled using the timestamps in the full data of \rsubsec{Prep}. The generated trajectory is further re-sampled by the given re-sampling rate for the real usage in the experiment.

\bsec{The Attention Mechansim}{AttentionVis}

\begin{figure}[t]
\centering
\includegraphics[height=1.8 in]{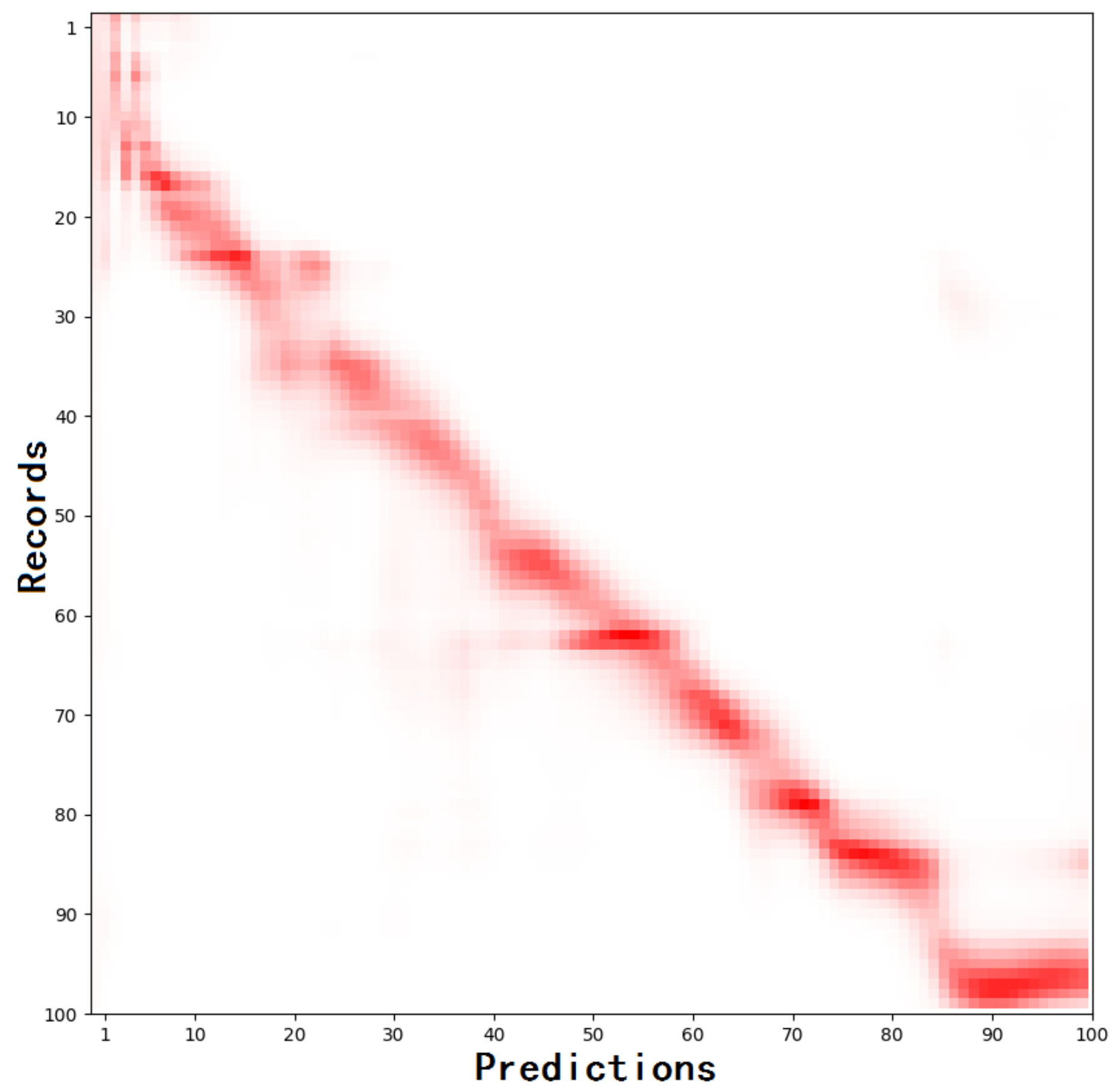}
\vspace{-0.07 in}
\caption{The attention matrix for a trajectory segment.}
\vspace{-0.13 in}
\label{fig:AttentionMatrix}
\end{figure}

\rfig{AttentionMatrix} visualizes the attention matrix $\textbf{A}$ of the encoder-decoder architecture computed by \req{Attention-2} in the mobility inference of one typical segment truncated from the trajectory ($L=100$). The nonzero values (red grids) happen close to the diagonal of the matrix, showing the local context used by the model. Initially at the beginning of the segment, the model requires a local context longer than 10 records for the cold start. After the model learns the global context of the trajectory, shorter local context is used.

\end{document}